\documentclass{article} 
\usepackage{iclr2026_conference,times}

\usepackage{amsmath,amsfonts,bm}

\def\eqref#1{equation~\ref{#1}}

\def\1{\bm{1}}

\DeclareMathAlphabet{\mathsfit}{\encodingdefault}{\sfdefault}{m}{sl}
\SetMathAlphabet{\mathsfit}{bold}{\encodingdefault}{\sfdefault}{bx}{n}

\usepackage{hyperref}
\usepackage{url}

\usepackage{wrapfig}
\usepackage{caption}

\usepackage[T1]{fontenc}    

\usepackage{enumitem}

\usepackage{graphicx}

\usepackage{algorithm}
\usepackage[noend]{algpseudocode}

\usepackage{booktabs}       
\usepackage{multirow}
\usepackage{colortbl} 

\usepackage[export]{adjustbox}

\title{DelvePO: Direction-Guided Self-Evolving \\ Framework for Flexible Prompt Optimization}

\author{Tao Tao, Guanghui Zhu\thanks{Corresponding Author}, Lang Guo, Hongyi Chen, Chunfeng Yuan, Yihua Huang\\
  State Key Laboratory for Novel Software Technology, Nanjing University\\
  taot@smail.nju.edu.cn, zgh@nju.edu.cn
}

\iclrfinalcopy 
\begin{document}

\maketitle

\begin{abstract}
Prompt Optimization has emerged as a crucial approach due to 
its capabilities in steering Large Language Models to solve 
various tasks. However, current works mainly rely on the random 
rewriting ability of LLMs, and the optimization process generally 
focus on specific influencing factors, which makes it easy to fall into local optimum. 
Besides, the performance of the optimized prompt is often unstable,
which limits its transferability in different tasks. 
To address the above challenges, we propose \textbf{DelvePO} 
(\textbf{D}irection-Guid\textbf{e}d Se\textbf{l}f-E\textbf{v}olving
Framework for Fl\textbf{e}xible \textbf{P}rompt \textbf{O}ptimization), 
a task-agnostic framework to optimize prompts 
in self-evolve manner. In our framework, we decouple 
prompts into different components that can be used to explore 
the impact that different factors may have on various tasks. 
On this basis, we introduce working memory, through which 
LLMs can alleviate the deficiencies caused by their own uncertainties 
and further obtain key insights to guide the generation of 
new prompts. Extensive experiments conducted on different 
tasks covering various domains for both open- and 
closed-source LLMs, including DeepSeek-R1-Distill-Llama-8B, Qwen2.5-7B-Instruct and GPT-4o-mini. Experimental results show that 
DelvePO consistently outperforms previous SOTA methods 
under identical experimental settings, demonstrating 
its effectiveness and transferability across different tasks.
\end{abstract}

\section{Introduction}
The rapid advancement of Large Language Models (LLMs) \citep{deepseekr1,searcho1} 
has revolutionized various real-world applications \citep{chinatravel,deepresearcher} . 
Prompt, a method that steers LLMs to produce desired results without 
modifying parameters, 
has garnered significant interest among non-AI experts 
from different domains \citep{wan2024teach,guo2025evopromptconnectingllmsevolutionary,fernando2023promptbreeder}. 
Consequently, the rapid growth in users has increased demand for prompt engineering methods. 

Previous efforts primarily focused on manually designing specialized prompts 
\citep{fewshotlearners,kojima2022large,wei2022chain}. 
However, this kind of method is time-consuming and 
demands extensive trial and error, making it less versatile 
for diverse tasks and limiting their real-world effectiveness. 
To reduce the human effort required for constructing effective prompts, 
many researches \citep{shum2023automatic,wang2022self,zhang2022automatic,feng2024auto,he2024does} have increasingly explored methods such as curating unified demonstrations 
for related tasks, systematically designing domain-specific templates, 
and identifying critical factors for prompt performance.
However, these methods exhibit limited applicability across diverse scenarios.

Subsequently,
a series of research emerged that employ optimization algorithms to refine prompts.
Such approaches (e.g. APE \citep{zhou2022large}, PromptBreeder \citep{fernando2023promptbreeder}, and EvoPrompt \citep{guo2025evopromptconnectingllmsevolutionary}) synergistically integrate the efficiency inherent in the algorithms 
with the powerful text processing ability of LLMs, 
achieving relatively stable performance enhancement on target datasets.
Although these studies analogize the mutation operation in 
evolutionary algorithms to the rewriting operation of LLMs,
they fail to fully harness the efficiency 
and rapid convergence inherent to such algorithms,
which ultimately limits the realization of 
their performance advantages in prompt optimization.
The primary reason lies in the inherently stochastic nature of 
the evolutionary process: 
the directionality of mutation operations remains uncontrollable, 
while their interpretability is also notably limited.
Furthermore, these methods neglect the potential impact of 
constituent components within a prompt on overall performance, 
leading to premature convergence in local optima.
For example, during evolutionary phase of EvoPrompt, 
the initial prompt inherently contains the "role" as a critical component. 
However, due to the stochastic nature of the mutation process, 
the stochastic mutation process may accidentally remove this component. 
Once discarded, it cannot be reintegrated into subsequent 
evolutionary iterations.
Such degradation significantly heightens the risk of premature convergence 
in local optima. A parallel limitation is observed in the PromptBreeder method,
which exhibits even higher stochasticity, as its implementation 
not only uses two distinct mutation prompts 
but also employs diverse mutation operators, 
amplifying randomness throughout the optimization process.
We summarize that a robust Prompt Optimization (PO) must have the following characteristics:

\begin{itemize}[leftmargin=*, labelindent=0pt, nosep, align=left]
    \item \textbf{Seamlessly integrating domain expert experience}: 
    For tasks in different domains, prior experience from domain experts can 
    be incorporated into the PO algorithm, 
    thus improving the efficiency of the optimization process.
    \item \textbf{Actively exploring factors that may 
    affect prompt performance}: The method can actively explore 
    factors affecting prompt performance to guide optimization using historical data.
    \item \textbf{Adaptively identifying optimal prompts 
    for different LLMs with varying performance}: The algorithm self-adjusts to discover the best prompts 
    for target tasks across differently specialized models 
    and scenarios, ensuring broad applicability in diverse 
    professional contexts.
  
\end{itemize}

Integrating insights from existing research,
we propose \textbf{DelvePO} \footnote{DelvePO is available at https://github.com/PasaLab/DelvePO} 
(\textbf{D}irection-Guid\textbf{e}d Se\textbf{l}f-E\textbf{v}olving
Framework for Fl\textbf{e}xible \textbf{P}rompt \textbf{O}ptimization) that adaptively accommodates diverse LLMs 
and self-improves through guidance from its historical 
optimization strategies. Inspired by the concept of Loci (the corresponding location of genes with important functions) and Alleles (different versions of the same gene) on genetics,  this framework first decouples prompt instructions into functional components (analogous to Loci). 
Subsequently, it iteratively evolves these 
components by exploring the potential impacts of diverse allele variations, ultimately achieving holistic optimization of the entire prompt through systematic recombination.
In particular, building upon the components, 
we introduce working memory mechanism (i.e., Component Memory and Prompt Memory) to guide the evolutionary process.
Component Memory is designed to capture evolutionary trends in 
individual components and utilize these trends to guide 
further optimization of each element. 
Take the component a step further, Prompt Memory creates interconnections between components by utilizing contextual information to guide the progressive optimization of the entire prompt.
The contributions of our work can be summarized as follows:\par
\begin{itemize}[leftmargin=*, labelindent=0pt, nosep, align=left]
    \item To the best of our knowledge, our work is the first to 
    introduce memory mechanism to guide prompt optimization,
    not only stabilizing the performance of the entire prompt population 
    but also greatly reducing the time required for evolutionary operations.
    \item By decoupling prompt into multiple components and designing guided evolutionary mechanisms, 
    our framework integrates multiple influencing factors 
    into a single prompt. This integration not only enhances 
    the scalability of PO methods but also improves the interpretability 
    of the optimization process, significantly lowering the difficulty to interact with the system.
    \item For LLMs with varying performance levels, 
    our framework can elicit their capabilities, 
    striking a good balance between exploring diverse components 
    and exploiting the current derived good components, 
    ultimately obtaining optimal prompts that  
    adapt to the target tasks and LLMs simultaneously. 
    Extensive experimental results
    on multiple 
    datasets and three widely-adopted LLMs reveal that DelvePO outperforms manually crafted prompts and existing PO methods.
\end{itemize}

\section{Preliminaries}

Given task $T=(\textbf{\emph{D}}, \textbf{\emph{A}})$, $\textbf{\emph{D}}$ is the task-related dataset and $\textbf{\emph{A}}$ represents the corresponding answer to the dataset, prompt optimization can be briefly described as follows:
Guided by the working memory mechanism, the initial prompt population $\textbf{\emph{P}}_{init}=\{p_1, p_2, \cdots \}$ is continuously optimized to obtain the final prompt population $\textbf{\emph{P}}_{final}$. The best prompt $p^*$ can be selected as follows: 

\begin{equation*}
p^* \leftarrow \mathop{\arg\max}\limits_{p \in {\textbf{\emph{P}}}_{final}}f_{eval}\big({\phi}^{\mathcal{LLM}}(p, \textbf{\emph{D}}_{dev}), \textbf{\emph{A}}\big)
\end{equation*}

where $\textbf{\emph{D}}_{dev}$ is the development dataset and ${\phi}^{\mathcal{LLM}}(p, \textbf{\emph{D}}_{dev})$ means that the prompts and questions are combined and then fed into the LLM to produce the corresponding response.
The important concepts used in our proposed framework are described below.

\textbf{Components} \ Similar to the relationship between Loci and Chromosome, components are mainly used to identify the location of key factors that affect task performance in prompts. Different tasks can introduce distinct components or reuse existing ones. 
Components are extensible, i.e., the type and number of components can be user-defined, and our method can also evolve synchronously as the context length that LLMs can receive increases. 
In this paper, we construct a comprehensive and representative component pool from a broad set of related literature.
Further details on how the components are studied and predefined in our framework are provided in Appendix \ref{app:B}.

\textbf{Templates} \ To bind components to prompts, we design a general template (corresponding to the Chromosome functionally), whose content is mainly composed of two parts: general and unchanging text; domain-specific and replaceable descriptive text (i.e., components and their corresponding values). For the descriptive text, its main functions include explaining domain-specific components, connecting different components, and providing contextual semantics about components. 
To overcome the instability of LLMs in recognizing components, we borrow the design idea of "markup" from HyperText Markup Language (HTML) to define different domain components. Taking "\textbf{\text{<}role\text{>}</role>}" as an example, the "role" is one of the various component types. Accordingly, the value of the component will be enclosed within the markup pairs, i.e., <role>\textbf{Sentence Simplifier}</role>.
More details can be found in Figure~\ref{fig:Template4Injection} in Appendix \ref{app:C}.

\section{Methodology}
\subsection{Framework of DelvePO}

\begin{figure}[t]
    \centering
    \includegraphics[width=1\textwidth]{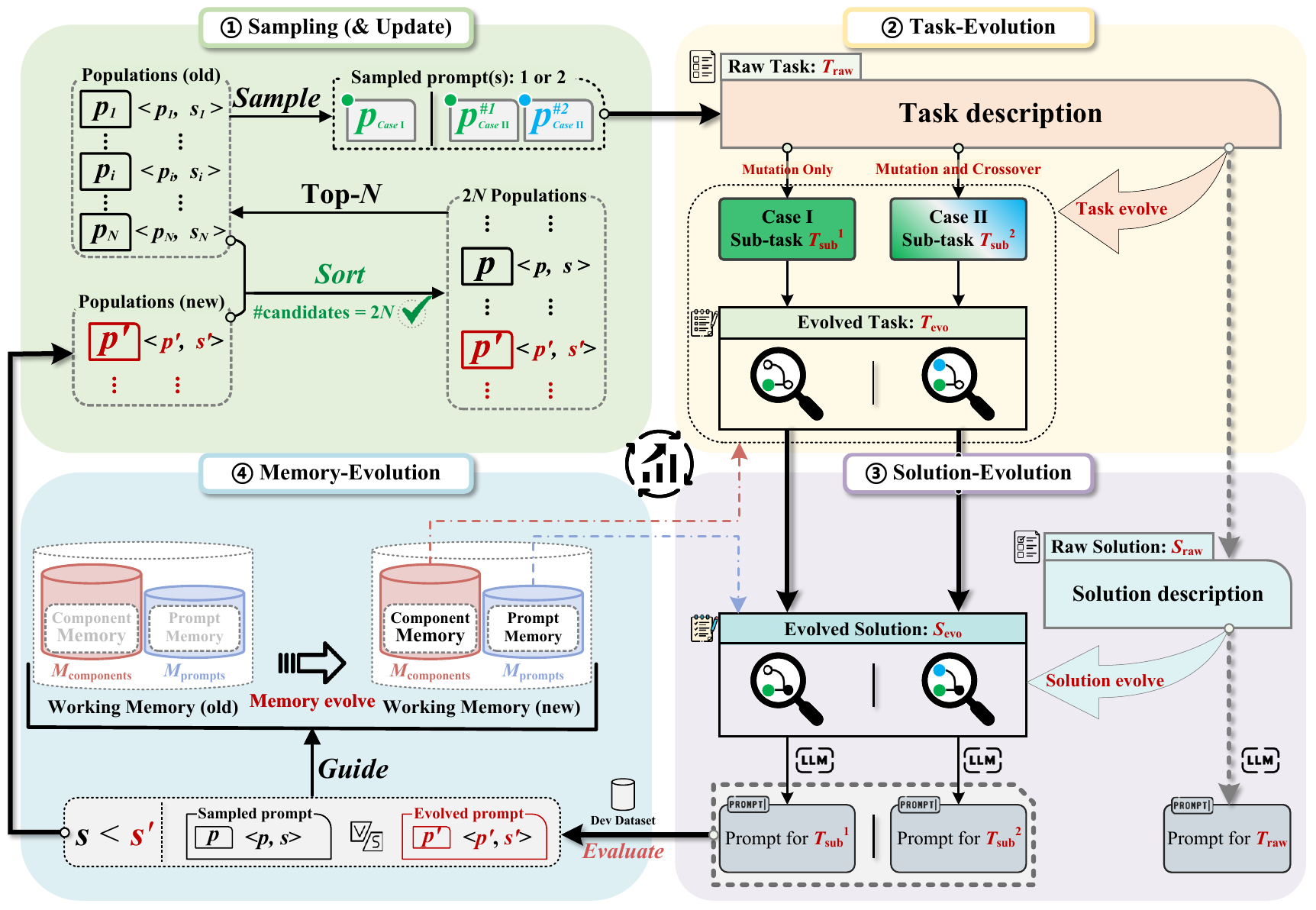}
    \caption{\textbf{The Framework of DelvePO}. Initialization begins with predefined components, which are concatenated to form individual $p$; multiple individuals constitute the initial population \textbf{Populations (old)}. At each step, one individual (Mutation only) or two individuals (Mutation and Crossover) are sampled, and the \textbf{Sub-task} determines the evolutionary direction (i.e., the mutated component type). Guided by \textbf{Task-}, \textbf{Solution-}, and \textbf{Memory-Evolution} modules, selected prompts are iteratively evolved, contrasting with unguided optimization. The new population \textbf{Populations (new)} is accumulated across epochs, and once the threshold is reached, the population is updated to initiate the next round of self-evolution.}
    \label{fig_framework}
    \vspace{-3ex}
\end{figure}

Our self-evolution prompt optimization framework consists of 4 necessary functional modules: Sampling \& Update module, Task-Evolution module, Solution-Evolution module and Memory-evolution module. We define the \textbf{Task} as "discover the promising direction of evolution", that is, determining the component (types or values) that need to evolve in the next step under the guidance of components memory. We define the \textbf{Solution} as "make sure the process of evolutionary operation and perform evolutionary operation", i.e., under the guidance of prompts memory, evolutionary operations are applied to the component values according to the selected evolution type: for a single sample, only mutation is performed, while for two samples, both mutation and crossover are executed. For memory-evolution, it mainly uses the evolved prompts and component value pairs before and after evolution to update the prompts memory and components memory, respectively. In the sampling and update module, when the number of iterations reaches a pre-defined value, the population is updated. Otherwise, a new sampling operation is performed within the current population, which in turn triggers the next round of self-evolution operations. The designs of \textbf{DelvePO} framework is shown in Figure \ref{fig_framework}.
Next, we first introduce the working memory mechanism.

\textbf{Components Memory} stores the corresponding component values before and after evolution, which is selected according to the mutated component type. The value pairs will be ordered by descend, i.e., when injecting to the final prompt, the first value performs better than the second. Components Memory will guide the selection of components in the Task-Evolution stage. 

\textbf{Prompts Memory} stores the prompts after each step of evolution. The evolved prompts are stored in descending order according to their performance scores. There are two forms of prompts memory: discrete form and continuous form. The discrete version only stores discrete combinations of component value in the prompt. And the continuous version stores a complete prompt formed by injecting component value into the template, which means that it stores continuous text containing context. Prompts memory will be used to guide the mutation of component or the crossover of the prompt in the Solution-Evolution stage. 

\subsection{Overview of DelvePO}

As shown in Figure \ref{fig_framework}, the workflow of \textbf{DelvePO} contains several core stages as outlined below.

\textbf{Initialization \& Sampling}: First, we use task-agnostic component-value generation prompt (see Figure~\ref{fig:Template4Component} in Appendix \ref{app:A}) to generate candidate values for each component type. Then, we randomly sample from these candidates and inject the selected values into the population-construction template (illustrated in Figure~\ref{fig:Template4Injection} in Appendix \ref{app:C}) to construct the initial population. 
Each individual in the initial population is evaluated on the development dataset to obtain its performance score. Finally, the sorted population is stored as the initial prompts memory. Before the population evolves, there is no components memory. After initialization, the sampling process begins, aiming to select prompts from the current population for evolution. Inspired by genetic principles, there are two main ways to generate new individuals: mutating a single individual or performing crossover between two individuals. Notably, mutation may also occur during crossover. To account for these cases, we assume that the number of individuals selected in each sampling step can be either 1 or 2.

\begin{figure}[htbp]
    \centering
    \includegraphics[width=1\textwidth]{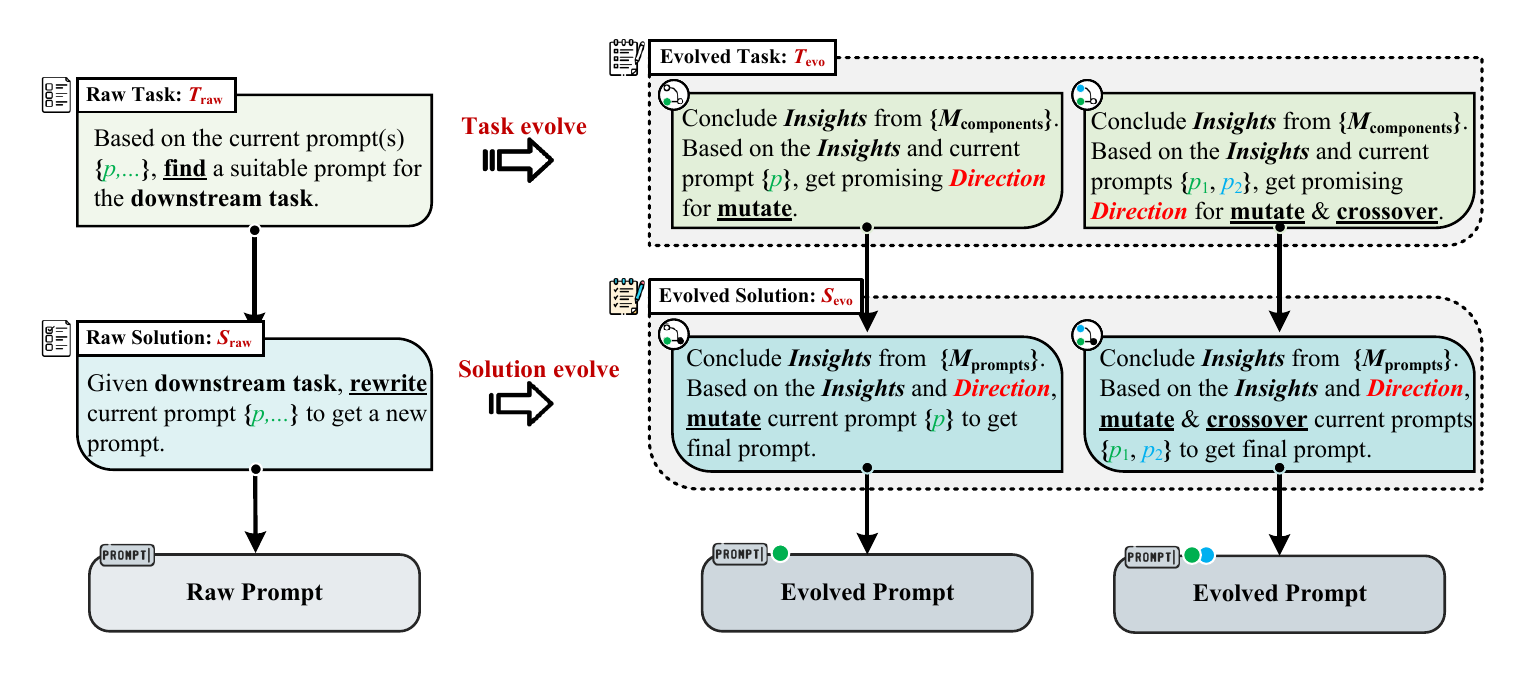}
    \caption{The mechanism of Task-Evolution and Solution-Evolution. \emph{Using the pseudo-prompt to explain the details of Task- and Solution-Evolution.}}
    \label{fig_evo}
    \vspace{-3ex}
\end{figure}

The evolutionary process mainly includes two parts: generating new individuals based on selected individuals; generating and storing the working memory. Specifically, there are 3 types of evolution, namely \textbf{Task-Evolution}, \textbf{Solution-Evolution}, \textbf{Memory-Evolution}. The mechanism of Task-Evolution and Solution-Evolution is shown in Figure \ref{fig_evo}.

\textbf{Task-Evolution} For task evolution, considering the components and the evolutionary operations (mutation and crossover), we design two kinds of evolutionary sub-tasks. The detailed information is shown in Figure~\ref{fig:subtask1_prompt} and Figure~\ref{fig:subtask2_prompt} (see Appendix \ref{app:D}).

\begin{itemize}[leftmargin=*, labelindent=0pt, nosep, align=left]

\item \textbf{Sub-task \uppercase\expandafter{\romannumeral1}}: This task mainly uses mutation operations to process a single candidate prompt. First, the semantic comprehension capability of the LLMs is utilized to obtain the relevant insights of component evolution from the component memory ${\textbf{\emph{M}}_\mathrm{components}}$. Then, the insights are used to guide the selection of components. Finally, the selected components will be treated as the promising direction to guide the evolution of mutation-based solution. 
\item \textbf{Sub-task \uppercase\expandafter{\romannumeral2}}: After performing Sub-task \uppercase\expandafter{\romannumeral1} on the two candidate prompts, we can get the respective component types set $C_1$ and $C_2$ for two prompts (say $p_{1}$ and $p_{2}$) as the promising direction for mutation. 
The final mutated component type is selected as $\hat{C}  = C_1 \cap C_2$. Next, for each component in $\tilde{C} = C \setminus \hat{C}$ where $C$ denotes the set of all component types, corresponding contents from $p_{1}$ and $p_{2}$ are extracted to construct a pair. Then, based on the insights derived from ${\textbf{\emph{M}}_\mathrm{components}}$, one value from each pair is selected as the potential value to improve performance of the prompts after evolution. Finally, the component types from $\hat{C}$ will be treated as the promising direction to guide the evolution of crossover-based solution, and the selected values from $p_{1}$ or $p_{2}$ whose component types coming from $\tilde{C}$ will also be passed into the corresponding Solution-Evolution phase to help construct the final prompts.
\end{itemize}

\textbf{Solution-Evolution} The main goal of solution evolution is to utilize the insights (derived from the prompts memory) and direction (received from the task-evolution) to perform evolution operations on the corresponding content in the current prompt and generate a new prompt that performs better. In this phase, we propose 2 sub-solutions corresponding to 2 sub-tasks. Depending on whether the prompt is continuous or discrete, each sub-solution can also be further divided to eliminate the effect of prompt format on the final result. 
\begin{itemize}[leftmargin=*, labelindent=0pt, nosep, align=left]
\item \textbf{Sub-solution \uppercase\expandafter{\romannumeral1}}: Extract component contents from current prompt based on the results obtained by sub-task \uppercase\expandafter{\romannumeral1} (i.e., the mutated components that are most likely to improve prompt performance). The extracted contents are then mutated using insights obtained from the prompts memory ${\textbf{\emph{M}}_\mathrm{prompts}}$ stored in discrete or continuous forms. 
Those contents that have not been mutated will be retained in new prompts. Finally, the mutated and unmutated component contents will be integrated as the result of sub-solution \uppercase\expandafter{\romannumeral1}. The corresponding prompts are shown in Figure~\ref{sc1}, \ref{sc2} (see Appendix \ref{app:E}) for the prompts memory in discrete and continuous forms, respectively. 

\item \textbf{Sub-solution \uppercase\expandafter{\romannumeral2}}: This mainly uses the results from sub-task \uppercase\expandafter{\romannumeral2} as a guide, and extracts component contents from the currently selected two prompts. And the evolutionary operations would combine mutation and crossover. First, for components that do not require mutation, the corresponding content is received from sub-task \uppercase\expandafter{\romannumeral2}. Then, for the component that need to be mutated, we extract its content from the two prompts. Based on the evolutionary insights derived from the prompt memory ${\textbf{\emph{M}}_\mathrm{prompts}}$, the mutation operations are performed on the extracted content. Next, the generated two prompts will crossover on the component types that need to be mutated. Finally, the results obtained from the mutation and crossover operations are integrated to generate a new prompt as the result of the sub-solution \uppercase\expandafter{\romannumeral2}. The details are shown in Figure~\ref{sc3} for the prompts memory in discrete form and Figure~\ref{sc41}, \ref{sc42} for continuous form (see Appendix \ref{app:E}).
\end{itemize}

\textbf{Memory-Evolution} is based on the component pairs and prompts both before and after the evolution to update the corresponding components memory and prompts memory, which is used to guide the next evolution process. In this module, the \textbf{evaluation} will be performed. Specifically, to clearly describe the evaluation process, we illustrate a general form of a prompt designed for LLMs that can be applied across different tasks (shown in Figure~\ref{Prompt2LLMs}). Evaluation refers to calculating the performance score of the generated new prompts on the development dataset based on the evaluation metrics of the target task, according to which components can be sorted and memory can be updated.

\textbf{Update}: Add the evolved prompts to the temporary population generated in each iteration. When the iteration ends, the temporary and current populations are mixed, and Top-N is selected as the updated population for the next iteration based on performance.

The details of \textbf{DelvePO} are outlined in Algorithm \ref{alg1}, which can be found in Appendix \ref{app:A}.

\section{Experiments}

\subsection{Experimental Settings}

\textbf{Baselines} In our experiments, We choose 6 commonly used methods which have been widely proven to be efficient in the field of prompt optimization as our baselines, which are: Crafted by human experts, CoT-ZS, CoT-FS, Promptbreeder, APE, and EvoPrompt.

\begin{itemize}[leftmargin=*, labelindent=0pt, nosep, align=left]
    \item \textbf{Human} corresponds to manually crafted prompts by experts, as detailed in the relevant literature \citet{zhang-etal-2024-sentiment, sanh2022multitaskpromptedtrainingenables}, which primarily derived from prior studies.
    \item \textbf{CoT} has been extensively applied in various domains, represents a rationale-based approach. We evaluate two representative forms of CoT: \textbf{CoT-ZS} (Zero-Shot CoT, \citet{kojima2022large}) and \textbf{CoT-FS} (Few-Shot CoT, also known as Manual-CoT, \citet{wei2022chain}).
    \item \textbf{APE} \citep{zhou2022large} regards instructions as programs and uses Monte Carlo Search to select appropriate instructions as optimized prompts under LLM guidance.
    \item \textbf{Promptbreeder} \citep{fernando2023promptbreeder} further investigates the effect of different mutation strategies on self-optimization based on elaborately designed evolutionary operations.
    \item \textbf{EvoPrompt} \citep{guo2025evopromptconnectingllmsevolutionary} introduces evolutionary algorithms to prompt optimization for the first time. Considering different scenarios, it instantiates its framework using two practical evolutionary algorithms. According to its statement, compared with GA method, the DE method has a wider range of use in solving complex problems. Therefore, we select EvoPrompt-DE as our baseline, and denote it simply as EvoPrompt.
\end{itemize}

\textbf{Datasets and LLMs} To demonstrate the generalizability of our method, we conducted experiments on 11 datasets across three LLMs, covering diverse domains and representative real-world tasks. The details information about datasets and LLMs are represented in Appendix \ref{app:dataset_and_llms}. Other experimental details (e.g., Computational Resources and Hyperparameter Details) are represented in Section~\ref{sec:repro}.

\subsection{Main Results}

Following the same settings as baselines, we tested the best prompts obtained during training. The main experimental results (as shown in Table~\ref{tab:ds_main}) on DeepSeek-R1-Distill-Llama-8B are reported as averages over three random seeds, with standard deviations provided. It is worth noting that we observed Promptbreeder to be significantly more time-consuming than other methods (as shown in Figure~\ref{timecost_main}). To balance the diversity of baselines and ensure the fairness in training time, we therefore report results for Promptbreeder using a single random seed.

From Table~\ref{tab:ds_main}, we can observe that our method achieves substantial improvements over manual approaches. Among the automated optimization methods, our method consistently outperforms baselines, demonstrating not only its effectiveness but also its adaptability to different task types. From the results on classical NLP benchmarks, we observe that the baselines perform well, confirming their effectiveness on established datasets. However, on more recently introduced benchmarks that demand broader capabilities, automated prompt optimization methods generally perform better, with our approach showing particularly substantial improvements. These results indicate that as LLMs continue to advance, prompt optimization techniques must likewise evolve, and our framework delivers consistently strong performance across diverse domains.

\begin{table}[ht]
\small
\centering
\caption{Main results on different downstream tasks for DeepSeek-R1-Distill-Llama-8B. Since expert-written prompts are not available for all datasets, sign ("-") is used to indicate missing cases.}
\label{tab:ds_main}
\resizebox{\textwidth}{!}{
\begin{tabular}{lllllllllc}
\toprule

\multirow{2}{*}{\textbf{Method}}  &
\multicolumn{3}{c}{\textbf{Classical NLP}} & 
\multicolumn{2}{c}{\textbf{Question-Answering}} &
\textbf{Domain-specific} &
\textbf{NLG} &
\multirow{2}{*}{\textbf{Avg.}} \\
\cmidrule(lr){2-4}
\cmidrule(lr){5-6}
\cmidrule(lr){7-7}
\cmidrule(lr){8-8}
&
\textbf{Subj} & \textbf{MR} & \textbf{CoLA} & \textbf{SQuAD} & \textbf{TREC} & \textbf{FinPB} & \textbf{SAMSum}\\

\midrule
Human & 26.00 & 55.89 & \text{-} & \text{-} & 54.67 & \text{-} & 25.68 & \text{-}\\

CoT-ZS & 70.00 & 68.00 & 65.45 & 43.91 & 68.00 & 60.00  & 3.23 & 56.74 \\

CoT-FS & \underline{83.00} & \underline{90.67} & \underline{70.63} & 47.92 & \underline{71.00} & 68.67 & 4.25 & 62.81 \\
\midrule

Promptbreeder & 35.00 & 86.00 & 55.58 & 54.16 & 60.00 & 59.00 & 27.88 & 51.20 \\

APE & 74.67$_{\textcolor{gray}{(2.85)}}$ & 83.67$_{\textcolor{gray}{(1.67)}}$ & 68.75$_{\textcolor{gray}{(1.20)}}$ & 67.57$_{\textcolor{gray}{(1.62)}}$ & 42.33$_{\textcolor{gray}{(2.40)}}$ & 70.67$_{\textcolor{gray}{(2.33)}}$ & \underline{30.02}$_{\textcolor{gray}{(0.85)}}$ & 61.25 \\

EvoPrompt & 82.00$_{\textcolor{gray}{(2.08)}}$ & 83.00$_{\textcolor{gray}{(1.00)}}$ & 66.75$_{\textcolor{gray}{(2.73)}}$ & \underline{68.17}$_{\textcolor{gray}{(1.14)}}$ & 67.00$_{\textcolor{gray}{(1.53)}}$ & \underline{72.00}$_{\textcolor{gray}{(1.53)}}$ & 29.18$_{\textcolor{gray}{(0.47)}}$ & \underline{65.55} \\

\midrule
\rowcolor{lightgray}

\textbf{DelvePO} & \textbf{83.67$_{\textcolor{gray}{(1.20)}}$} & \textbf{91.00$_{\textcolor{gray}{(1.00)}}$} & \textbf{76.25$_{\textcolor{gray}{(1.49)}}$} & \textbf{68.53$_{\textcolor{gray}{(2.61)}}$} & \textbf{76.00}$_{\textcolor{gray}{(2.08)}}$  & \textbf{73.33$_{\textcolor{gray}{(3.06)}}$} & \textbf{32.05$_{\textcolor{gray}{(0.25)}}$} & \textbf{70.48} \\

\bottomrule
\end{tabular}
}
\vspace{-3pt}
\end{table}

To further evaluate the performance of our framework on different LLMs, we conducted additional experiments across different task types on the closed-source model (GPT-4o-mini, results reported in Table~\ref{tab:gpt_4o_mini_main}) and the widely used open-source model (Qwen2.5-7B-Instruct, shown in Table~\ref{tab:qw} in Appendix \ref{app:exp_more}). The experimental settings were kept identical to the main experiments. As shown in the results evaluated on these two LLMs, our framework consistently delivers either superior or competitive performance across multiple task types, demonstrating its robustness and general effectiveness when applied to diverse LLMs.

\begin{table}[t]
\small
\centering
\caption{The results on different downstream tasks for GPT-4o-mini.}
\label{tab:gpt_4o_mini_main}
\begin{center}
\begin{tabular}{cllllc}
\toprule
\multirow{2}{*}{\textbf{Method}}  &
\multicolumn{2}{c}{\textbf{Classical NLP}} & 
\textbf{Domain-specific} &
\textbf{Multi-domain} &
\multirow{2}{*}{\textbf{Avg.}} \\
\cmidrule(lr){2-3}
\cmidrule(lr){4-4}
\cmidrule(lr){5-5} &
\textbf{Subj} & \textbf{CoLA} & \textbf{FinPB} & \textbf{AG's News} \\

\midrule
Human & 27.33 & \text{-}  & \text{-} & \underline{87.56}& \text{57.45} \\

CoT-ZS & 67.67 & 81.40 & 73.67 & 80.33 & 75.77 \\

CoT-FS & 82.00 & \textbf{84.93} & 80.67 & 83.00 & 82.65 \\

\midrule
Promptbreeder & 45.00 & 67.72 & 72.00 & 78.00 & 65.68\\

APE & \underline{79.61}$_{\textcolor{gray}{(1.78)}}$ & 81.53$_{\textcolor{gray}{(1.93)}}$ & 94.93$_{\textcolor{gray}{(0.78)}}$ & 84.60$_{\textcolor{gray}{(0.93)}}$ & 85.17  \\

EvoPrompt & 76.70$_{\textcolor{gray}{(1.90)}}$ & 82.72$_{\textcolor{gray}{(2.11)}}$ & \underline{96.97}$_{\textcolor{gray}{(0.52)}}$ & 86.50$_{\textcolor{gray}{(1.40)}}$ & \underline{85.72} \\
\midrule
\rowcolor{lightgray}
\textbf{DelvePO} & \textbf{91.07$_{\textcolor{gray}{(1.03)}}$} & \underline{83.14}$_{\textcolor{gray}{(1.90)}}$ & \textbf{98.63$_{\textcolor{gray}{(0.62)}}$} & \textbf{89.40$_{\textcolor{gray}{(0.81)}}$} & \textbf{90.56}  \\
\bottomrule
\end{tabular}
\end{center}
\vspace{-4ex} 
\end{table}

\subsection{Cost Analysis} 
\begin{figure}[htbp]
    \centering
    \includegraphics[width=0.8\textwidth]{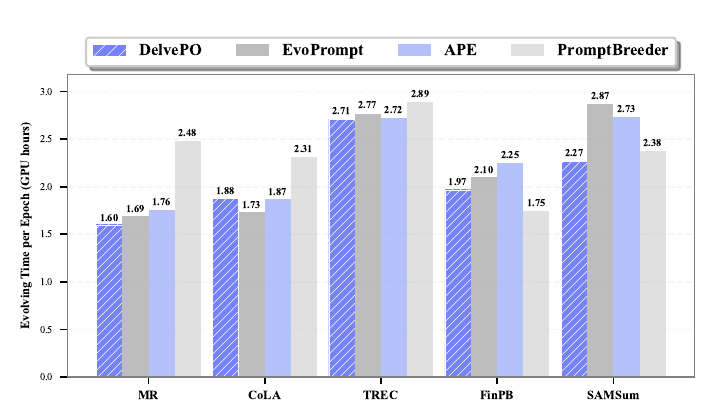}
    \captionsetup{skip=1pt} 
    \caption{Average time-consuming (GPU hours) for one epoch of optimization on DeepSeek-R1-Distill-Llama-8B.}
    \vspace{-4pt} 
    \label{timecost_main}
\end{figure}

In our experiments, the overhead primarily stems from the training time required for open-source LLMs and the number of tokens consumed in API requests for closed-source LLMs. Accordingly, for DeepSeek-R1-Distill-Llama-8B, we randomly selected one dataset from each task type and measured the time cost of different baselines, with results presented in Figure \ref{timecost_main}. The statistics indicate that our method consistently outperforms or matches the baselines in terms of optimization speed, particularly when compared with PromptBreeder. This also explains why we report its results using a single random seed. Overall, the results demonstrate that our method can more effectively exploit the rapid convergence property of evolutionary algorithms for faster optimization.

Moreover, we reported token usage in terms of the actual monetary expenditure, as shown in Table~\ref{tab:gpt4omini_tokens}. Overall, as shown in Table~\ref{tab:gpt_4o_mini_main} and Table~\ref{tab:gpt4omini_tokens}, although our method requires higher expenditure, it consistently delivers performance above or competitive with the baselines, indicating that our approach offers a favorable balance between performance and cost. We also analyzed the reasons behind the generally higher token usage. The primary factor is that the content stored in the memory module is included as part of the input provided to the target LLMs. In future work, we plan to integrate prompt compression techniques into the framework to reduce this overhead.

\subsection{Ablation Study}

To evaluate the impact of the memory mechanism in our framework, we conducted ablation experiments on GPT-4o-mini. We selected three datasets of different types to evaluate the adaptability of the memory mechanisms across multiple scenarios. 
Table~\ref{tab:ablation_memory} reports the performance on three types of datasets using a single random seed. 
%
%
When both memory mechanisms are included and operate in coordination, the overall performance is substantially higher than in the other configurations, demonstrating the effectiveness and complementary benefits of the proposed memory design.

\begin{table}
\small
\centering
\caption{Ablations of Memory Mechanism.}
\label{tab:ablation_memory}
\begin{tabular}{lcccc}
\toprule
\textbf{Memory Modules} & \textbf{SAMSum} & \textbf{SQuAD} & \textbf{Causal Judgement}\\
\midrule
w/o Component Memory & 28.8 & 67.4 & 62.6\\
w/o Prompt Memory & 29.4 & 67.9 & 61.8\\
w/o both & 28.4 & 64.6 & 61.3 \\
\textbf{DelvePO} & \textbf{35.3} & \textbf{84.7} & \textbf{65.7}\\
\bottomrule
\end{tabular}
\vspace{-4pt}
\end{table}

\begin{wraptable}{r}{0.5\textwidth}
\vspace{-2pt}     
\centering
\caption{Sensitivity test regarding the number of component values.}
\label{tab:component_number_sensitivity}
\begin{tabular}{lccc}
\toprule
\textbf{\# Value} & \textbf{SAMSum} & \textbf{SQuAD} & \textbf{SST-5} \\
\midrule
50 & 29.2 & 67.9 & 57.2 \\
40 & 29.2 & 67.3 & 57.4 \\
30 & 29.7 & 66.8 & 56.8 \\
20 & 28.8 & 66.5 & 59.1 \\
10 & \textbf{30.2} & \textbf{69.7} & \textbf{60.3} \\
\bottomrule
\end{tabular}
\vspace{-4pt} 
\end{wraptable}

Furthermore, to investigate the impact of the number of component values for each component type on the overall performance of initial population, we designed a sensitivity test examining how initial population performance varies with the number of component values at initialization. Using GPT-4o-mini, we generated initial populations for three different types of datasets under a single random seed and evaluated their performance. The results in Table~\ref{tab:component_number_sensitivity} show that increasing the number of component values does not cause significant fluctuations in the initial population performance. This indicates that a relatively small number of component values is sufficient to obtain an initial population with stable and reasonable performance, and importantly, it rules out the concern that a larger number of components could lead to an overestimated initial population, which might otherwise suggest that further optimization is unnecessary. 


Moreover, we conducted a \textbf{case study} to help researchers quickly understand the operational
mechanisms of our proposed framework. The details are presented in Appendix~\ref{app:case_study}.

\section{Related Work}


\textbf{Prompt Engineering} \ Prompt engineering is a resource-efficient approach, focusing on elaborately 
designing expert-level prompts to steer LLMs generate desired solutions to 
various downstream tasks. In this part, we mainly focus on those works 
which use prompts to stimulate the internal abilities of LLMs. Least-to-Most \citep{zhou2022least},
Decomposed Prompting \citep{khot2022decomposed} and PS\&PS+ \citep{wang2023plan} 
use prompts to leverage the decomposition ability of LLMs, 
breaking down complex problems into simpler ones, 
enabling the model to perform better when dealing with complex problems.
CoT \citep{wei2022chain}, PoT \citep{chen2022program}, PS \& PS+ \citep{wang2023plan}, 
Automate-CoT \citep{shum2023automatic}, ToT \citep{yao2023tree}and 
GoT \citep{besta2024graph} guide the model to utilize 
chain-of-thought in different ways through the design of prompts, 
stimulating the thinking ability of the model, thereby enhancing 
the model's reasoning ability. 
Also, Complexity-based Prompting \citep{fu2022complexity}
and DIV-SE \citep{naik2023diversity} focus on the complexity and diversity 
of prompts, aiming to help the model think better.
Rephrase and Respond \citep{deng2023rephrase}, OPRO \citep{yang2023large}, 
and MIPRO \citep{opsahl2024optimizing} utilized the 
self-optimization capabilities of LLMs through methods such as 
input rewriting, iterative prompt optimization and structured program 
optimization, jointly demonstrating that LLMs can autonomously 
enhance the performance of task execution by dynamically improving prompts.
TextGrad \citep{yuksekgonul2024textgrad} and SPO \citep{xiang2025self} combine LLMs by orchestrating 
Standard Operation Pipelines (SOPs) in advance, and uses the evaluation ability of the model 
itself to guide the optimization of prompts. 
These methods effectively demonstrate that LLMs can be more proactive 
in utilizing their exploration abilities under the scientific guidance 
of predefined SOPs.
Although the above works have elicited some abilities of LLMs to cope with complex 
problems, they cannot get rid of the problem that LLMs 
are sensitive to inputs, which results in the inconsistency of outputs' quality.

\textbf{Prompt Optimization} \ 
Given a downstream task, prompt optimization aims to improve 
the effectiveness of prompt, which typically involves an 
iterative process including initialization, execution, evaluation 
and selection. This part primarily focus on those works which leverage external technologies or exogenous intelligence sources to guide LLMs to perform prompt optimization. Using external knowledge to optimize  prompt is very effective. Existing works generally 
referred to: 1) the way humans think \citep{wang2022self}; 2) the idea of program synthesis \citep{zhou2022large}; 3) external knowledge \citep{zhao2023verify} to optimize prompts which achieve good results.
Formatting the structure of prompts can standardize 
the thinking process of LLMs, and to a certain extent improve 
their reasoning capability. LangGPT \citep{wang2024langgpt} presents a framework 
for prompt design, proving that scalable structures 
are important for prompts migration. Prompt template \citep{he2024does}
delves into the impact of the format of the prompt template on solving problems, demonstrating the effectiveness 
of structured prompts in eliciting LLMs' capabilities. 
Furthermore, there are some efforts that introduce algorithms 
that have been widely proven to have good optimization capabilities 
to the optimization of prompts, including K-means \citep{zhang2022automatic}, 
KNN \citep{shi2022knn}, reinforcement learning \citep{pryzant2023automatic,wang2023promptagent}, 
active learning \citep{diao2023active}, and evolutionary algorithm \citep{guo2025evopromptconnectingllmsevolutionary,fernando2023promptbreeder}.

In summary, although existing studies have mitigated the output stochasticity of LLMs, the 
efficiency of the optimization algorithm has still not been fully explored. These efforts generally tend to treat prompts as a whole unit to optimize, so the potential optimization space is very large. 
In addition, most previous researches combining optimizing algorithms (e.g., evolutionary algorithms) with LLMs, 
do not take full advantage of the experience generated before and after optimization, so that the optimization 
process is more stochastic, which tends to fall into local optima. Inspired by biological Loci 
and Alleles, 
this paper proposes a flexible framework for prompt optimization, which can effectively reduce the randomness 
of the optimization process and significantly improves the optimization speed. We hope our approach will provide possible improvements 
for subsequent PO methods, significantly lowering the learning barrier for non-AI experts to leverage LLMs.

\section{Conclusion}
We introduced DelvePO, a self-evolving framework for prompt optimization that decouples prompts into distinct components. With components, prompts can be modified by adding or removing content that may affect their performance, striking a good balance between exploration and exploitation of factors that affect task performance. DelvePO employs a co-evolutionary mechanism to iteratively refine the specifics of  two sub-tasks and generate corresponding solutions. The evolved prompt, following systematic processing, is encoded into working memory to facilitate LLMs in deriving relevant insights, thereby provides directional guidance for generating task-specific prompts. Extensive experiments on different tasks demonstrate DelvePO consistently outperforms baselines, validating its effectiveness. As we anticipate the emergence of even more powerful LLMs that can deal with longer context, we firmly believe that more professional prompts will penetrate all walks of life, and DelvePO will help more users complete various complex tasks.

\clearpage

\section*{Ethics Statement}

This work studies prompt optimization techniques for language models (LLMs) 
to better elicit their capabilities in solving target tasks. The primary potential 
risks of this research are related to the misuse of LLMs, for example, generating 
misleading, harmful, or biased content. 

In our experiments, we only use publicly available datasets and pre-trained LLMs, 
and no private or sensitive data were involved. Specific statements on LLM usage can be found in Appendix \ref{app:use_of_LLMs}. We emphasize that our methods are intended for research and benchmarking purposes, 
and we encourage responsible use to mitigate potential societal risks.

\section*{Reprodicibility Statement}
\label{sec:repro}

We are committed to ensuring the reproducibility of our work. To facilitate replication, 
we provide the following details:

\textbf{Computational Resources} The following describes the experimental environment, including detailed information on both hardware and software configurations.	
\begin{itemize}[leftmargin=*, labelindent=0pt, nosep, align=left]
    \item \textbf{Hardware}. All experiments were conducted on a computing node equipped with four NVIDIA Tesla V100-SXM2 GPUs (32GB memory each), an Intel Xeon Gold 6248 CPU @ 2.50GHz with 20 cores, and 226 GB of RAM. 
	\item \textbf{Software}. The system runs Ubuntu 20.04.6 LTS with Linux kernel version 5.4.0. All models were implemented in Python 3.10.18 using PyTorch 2.0.0 with CUDA 11.7. 
\end{itemize}

\textbf{Hyperparameter Details} In order to isolate the effect of our proposed method and ensure a fair comparison, we mainly followed the default configurations used in baseline methods and intentionally introduced no additional trainable parameters. Specifically, the detailed hyperparameter settings are given below.

\begin{itemize}[leftmargin=*, labelindent=0pt, nosep, align=left]
    \item \textbf{Initial Population Size}. Following the setup of EvoPrompt, which uses both human-written and LLM-generated prompts, we adopted a similar strategy in spirit but tailored it to our fully automated framework. (1) We identify a fixed set of components through preliminary study mentioned at ref~\label{xxx}. (2) For each component, we use an LLM to generate 10 candidate values based on prompt templates. (3) We then randomly combine these values to create 10 initial prompts, which together form the initial population for the evolutionary process.

\item \textbf{Temperature}. Since the stochasticity of LLM outputs is sensitive to temperature settings, we set the temperature to 0.5 to strike a balance between exploration and exploitation. This choice aligns with prior work such as EvoPrompt.

\item \textbf{Sample Allocation}. For data splits, we followed the protocols of APE and EvoPrompt. Specifically, if the dataset has a predefined training/testing split, we used it as-is. For datasets without predefined splits, we randomly selected 100 examples as the test set and used the remaining examples for training.
\item \textbf{Randomness Control}. To ensure reproducibility. Unless otherwise noted, we use 3 random seeds (5, 10 and 15) in the training phrase, and reported the results on the test set. 
\end{itemize}

\section*{Limitations}
While our framework can adaptively design well-matched prompts for any LLM across diverse downstream tasks, several limitations remain. (1) Due to substantial computational costs, we cannot comprehensively evaluate all models and domains. Instead, we focused on widely used datasets to balance fairness and coverage. (2) Although we report monetary cost based on actual token usage, variations in token pricing across input and output types cannot be precisely captured by the API. Analysis indicates that most of the cost arises from including memory content as input tokens, while output token consumption remains relatively modest, particularly when "thinking mode" is disabled. Future work will explore prompt compression to further optimize resource use. (3) We evaluated only representative component values from each category due to resource constraints. Nevertheless, even with this limited set, our approach continues to outperforms or remains competitive with baselines, demonstrating its effectiveness and suggesting that its benefits will likely increase as LLMs support longer contexts.

\newpage

\bibliography{iclr2026_conference}
\bibliographystyle{iclr2026_conference}

\clearpage

\appendix

\section{Use of LLMs}
\label{app:use_of_LLMs}

Large Language Models (LLMs) were used in two ways in this work. First, LLMs served as 
base models in our experiments on prompt optimization, where we studied how different 
prompts can elicit their capabilities to solve target tasks. Second, LLMs were employed 
as auxiliary tools for minor writing support, such as grammar checking and phrasing 
improvements. Specific details about the LLMs used in our experiments can be found in 
Appendix~\ref{app:dataset_and_llms}. No LLMs were used to generate substantive ideas, analyses, or content of the paper.

\section{Details of Datasets and LLMs Used}
\label{app:dataset_and_llms}

\textbf{Datasets} For fair comparison, we followed the datasets and evaluation metrics used in prior baselines whenever possible. Specifically, we include 4 classic NLP benchmarks (\emph{MR}, \emph{Subj}, \emph{CoLA}, \emph{SST-5}) and two widely used question-answering datasets (\emph{SQuAD}, \emph{TREC}) to validate basic capabilities; several domain-specific benchmarks to probe specialized performance, including \emph{Financial Sentiment Evaluation} dataset (\emph{FinFE}), \emph{Financial PhraseBank} (\emph{FinPB}), reasoning related dataset (\emph{Casual Judgement}). Besides, one multi-domain datasets (\emph{AG’s News}) and one natural language generation dataset (\emph{SAMSum}) are also used to assess overall robustness. To evaluate output quality beyond simple accuracy, we report ROUGE-Avg on \emph{SAMSum} and the Matthews correlation coefficient (MCC) on \emph{CoLA}. To balance computational cost while maximizing coverage, we selected datasets according to a “maximize capability diversity” principle — for example, in addition to the main experiments we ran Qwen2.5-7B-Instruct on \emph{Subj}, \emph{AG’s News}, and \emph{FinFE} to cover several of the categories above. Detailed results are presented in the experimental analysis section.

\textbf{LLMs} To demonstrate the adaptability of the proposed method for LLMs, we selected \emph{DeepSeek-R1-Distill-Llama-8B} and \emph{Qwen2.5-7B-Instruct} from open-source LLMs, as well as \emph{GPT-4o-mini} from closed-source LLMs, as the base models for our experiments. The experiments on \emph{DeepSeek-R1-Distill-Llama-8B} evaluate both the performance of the DeepSeek model itself and, to some extent, the capabilities of the underlying Llama architecture, which is primarily trained on English-language data. Experiments on \emph{Qwen2.5-7B-Instruct} assess the framework's performance on a model predominantly trained on Chinese-language data, demonstrating applicability to non-English corpora. \emph{GPT-4o-mini} was included because it is a widely used closed-source model in prior studies and allows cost-effective experimentation within our budget.

\section{Algorithm Details}
\label{app:A}

\begin{algorithm}[htbp]
    
	\caption{An Overview of \textbf{DelvePO}}
	\label{alg1}
	\begin{algorithmic}[1]
        \Require A population of prompts $\textbf{\emph{P}}$, size of population $N$, task-related dataset $\textbf{\emph{D}}$, number of epochs $m$, number of iterations $n$, working memory $\textbf{\emph{M}} = \left\{\textbf{\emph{M}}_{\mathrm{components}},\textbf{\emph{M}}_{\mathrm{prompts}}\right\}$
        \Ensure Best prompt $p^*$
		\State \textbf{Initialization}: $\textbf{\emph{P}}=\left\{p_1, p_2, \cdots, p_N\right\}$, $\textbf{\emph{M}}_{\mathrm{prompts}} \leftarrow f_{sort}({\textbf{\emph{P}})} $, $\textbf{\emph{M}}_{\mathrm{components}} \leftarrow \emptyset$
        \For{epoch = 1 to $m$}
            \State $\textbf{\emph{P}}_{\mathrm{evo}} \leftarrow \emptyset$
            \For{step = 1 to $n$}
                \State \textbf{Selection}: $p \leftarrow f_{r.w.s.}(\textbf{\emph{P}})$
                \State \textbf{\underline{Task-Evolution}}: $\mathcal{T}_{\mathrm{evo}} \leftarrow {\phi}^{\mathcal{T}}(p,\textbf{\emph{M}}_{\mathrm{components}} \ |\ \mathcal{T})$  
                \State \textbf{\underline{Solution-Evolution}}: $\mathcal{S}_{\mathrm{evo}} \leftarrow  {\phi}^{\mathcal{S}}(p, \textbf{\emph{M}}_{\mathrm{prompts}}\ | \ \mathcal{T}_{\mathrm{evo}})$ 
                \State \textbf{Evaluation}: $p' \leftarrow {\phi}^{\mathcal{LLM}}(\mathcal{S}_{\mathrm{evo}})$, \ $s' \leftarrow f_{eval}(p', \textbf{\emph{D}})$ 
                \State \textbf{\underline{Memory-Evolution}}: $\textbf{\emph{M}}_\mathrm{evo} \leftarrow {\phi}^{\mathcal{M}}\big( \textbf{\emph{M}},  \langle p, p', s\geq s' \rangle\big)$ 
                \State $\textbf{\emph{P}}_{\mathrm{evo}} \leftarrow \left\{\textbf{\emph{P}}_{\mathrm{evo}}, p'\right\}$
            \EndFor                            
            \State {\textbf{end for}}
                \State \textbf{Update}: $\textbf{\emph{P}}  \leftarrow \text{Top-}N \left\{\textbf{\emph{P}},\textbf{\emph{P}}_{\mathrm{evo}}\right\} $
        \EndFor
        \State {\textbf{end for}}
        \State {\textbf{Return}} the best prompt $p^*$: $p^* \leftarrow \mathop{\arg\max}\limits_{p \in {\textbf{\emph{P}}}}f_{eval}\big({\phi}^{\mathcal{LLM}}(p, \textbf{\emph{D}})\big)$
	\end{algorithmic}  
\end{algorithm}

\begin{figure}[htbp]
    \centering
    \includegraphics[width=0.95\textwidth,cframe=black!50!black 0.3mm]{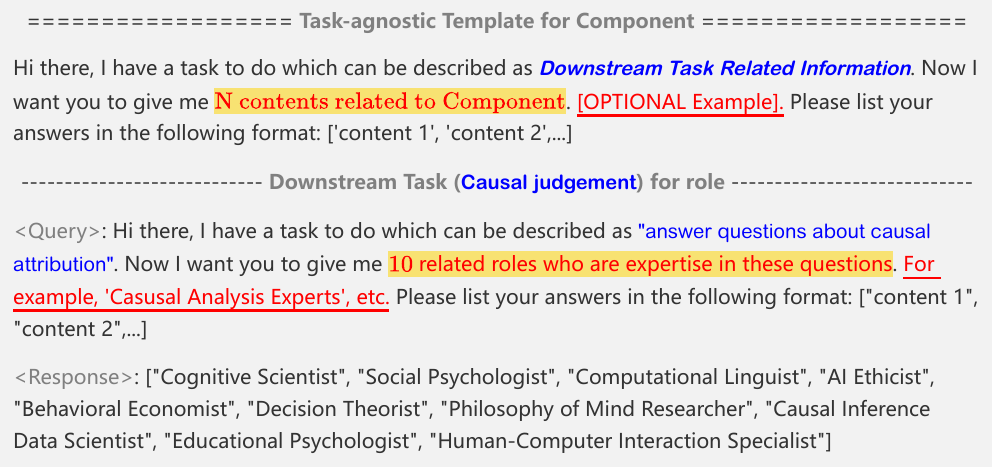}
    \caption{Task-agnostic template for generating component values corresponding to the given component types. \emph{The following part of the figure is the prompt to generate content for Component "role" using the casual judgement task as an example.}}
    \label{fig:Template4Component}
\end{figure}

The sampling function used in our framework is roulette wheel selection, denoted as $f_{r.w.s.}(\cdot)$, which is commonly used in the evolution algorithm. $\phi^{\mathcal{T}}$, $\phi^{\mathcal{S}}$, $\phi^{\mathcal{M}}$ refer to the Task-Evolution, Solution-Evolution, Memory-Evolution methods, respectively. Similarly, $\mathcal{T}$, $\mathcal{S}$, and $\mathcal{M}$ mean the corresponding Task, Solution, Memory. Based on the components, we designed a task-agnostic template described in Figure~\ref{fig:Template4Component}, through which any kind of LLMs can construct an initial content set of components based on a simple description of the target task input by the user. 

\section{Additional Experiments}
\label{app:exp_more}

\begin{table}[ht]
\small
\centering
\caption{The results on different downstream tasks for Qwen2.5-7B-Instruct.}
\label{tab:qw}
\resizebox{\textwidth}{!}{
\begin{tabular}{lccccccc}
\toprule
\multirow{2}{*}{\textbf{Method}}  &
\multicolumn{3}{c}{\textbf{Classical NLP}} &
\textbf{Question-Answering} &
\textbf{Domain-specific} &
\textbf{Multi-domain} &
\multirow{2}{*}{\textbf{Avg.}} \\
\cmidrule(lr){2-4}
\cmidrule(lr){5-5} 
\cmidrule(lr){6-6} 
\cmidrule(lr){7-7} 
&
\textbf{Subj} & \textbf{SST-5} & \textbf{CoLA} & \textbf{TREC} & \textbf{FinFE} & \textbf{AG's News} \\
\midrule
APE & 69.00$_{\textcolor{gray}{(3.06)}}$ & 47.00$_{\textcolor{gray}{(1.10)}}$ & 79.05$_{\textcolor{gray}{(1.73)}}$ & 43.40$_{\textcolor{gray}{(1.14)}}$ & 64.30$_{\textcolor{gray}{(2.70)}}$ & 83.43$_{\textcolor{gray}{(1.90)}}$ & 64.38 \\
EvoPrompt & \underline{77.03}$_{\textcolor{gray}{(4.74)}}$ & \underline{57.67}$_{\textcolor{gray}{(1.19)}}$ & \underline{79.69}$_{\textcolor{gray}{(1.42)}}$ & \underline{67.55}$_{\textcolor{gray}{(2.08)}}$ & \underline{64.67}$_{\textcolor{gray}{(1.22)}}$ & \underline{85.73}$_{\textcolor{gray}{(1.29)}}$ & \underline{72.06} \\
\midrule
\rowcolor{lightgray}
\textbf{DelvePO} & \textbf{80.07$_{\textcolor{gray}{(0.65)}}$} & \textbf{60.00$_{\textcolor{gray}{(1.69)}}$} & \textbf{81.40$_{\textcolor{gray}{(1.07)}}$} & \textbf{70.77$_{\textcolor{gray}{(1.74)}}$} & \textbf{69.97$_{\textcolor{gray}{(0.87)}}$} &  \textbf{89.27$_{\textcolor{gray}{(0.97)}}$}  & \textbf{75.25} \\
\bottomrule
\end{tabular}
}
\end{table}

\begin{table}[H]
\centering
\caption{Average monetary cost (USD) for one epoch of optimization on GPT-4o-mini.}
\label{tab:gpt4omini_tokens}
\begin{tabular}{lccccc}
\toprule
\textbf{Methods} & \textbf{Subj} & \textbf{CoLA} & \textbf{FinPB} & \textbf{AG's News}\\
\midrule
Promptbreeder & 1.17 & 1.31 & 0.97 & 1.52\\
APE & 0.57 & 0.56 & 0.61 & 0.79\\
EvoPrompt & 0.83 & 0.64 & 0.74 & 1.23\\
DelvePO & 1.27 & 1.08 & 1.30 & 1.10\\
\bottomrule
\end{tabular}
\end{table}

\clearpage

\section{Detailed Information about Components}
\label{app:B}

To ensure that the types of components are as comprehensive and representative as possible, we first surveyed a broad set of related literature \citep{yuksekgonul2024textgrad, he2024does, feng2024auto, opsahl2024optimizing, diao2023active, wang2024langgpt, wang2023promptagent} and extracted a variety of factors that have been shown to influence the performance of prompts, forming our component pool. We then categorized all components in the pool based on the semantics implied in their original sources, which resulted in five categories: “Role and Expertise”, “Task Content”, “Constraints and Norms”, “Process and Behavior” and “Context and Examples”. From each category, we selected the most representative component as our predefined component types. The complete component pool and its categorization are provided in Table~\ref{tab:component_pool}. 

Despite this extensive literature review, we acknowledge that some important aspects may remain uncovered. This observation motivated our design: as more non-AI experts begin to use LLMs, domain specialists should be able to adaptively define new components through our mechanism, thereby supporting both effective task performance and improved interpretability. It is worth noting that for each component type, we can add a “null” option when generating its values, allowing the presence or absence of the component to be controlled and makes the optimized prompts more flexible.

\begin{table}[ht]
\small
\centering
\caption{The categories and types of components in the component pool}
\label{tab:component_pool}
\begin{tabular}{ll}
\toprule
\textbf{Categories} & \textbf{Related Items} \\
\midrule
Role and Expertise & \underline{Role}; Role description; Scenario; Domain knowledge; Term Clarification \\

Task Content & \underline{Task description}; Instruction; Goal \\

Constraints and Norms & \underline{Output format}; Constraints; Principle; Style; Length; Tone; Priority \& \\
 & Emphasis; Exception handling; Target audience \\

Process and Behavior & \underline{Workflow}; CoT; Action; Skill; Suggestions; Initialization \\

Context and Examples & \underline{Examples}; Reference prompt; Attachment \\
\bottomrule
\end{tabular}
\end{table}

\section{Template for Injection \& Prompts for Evaluation on LLMs}
\label{app:C}

\begin{figure}[H]
    \centering
    \includegraphics[width=0.95\textwidth,cframe=black!50!black 0.3mm]{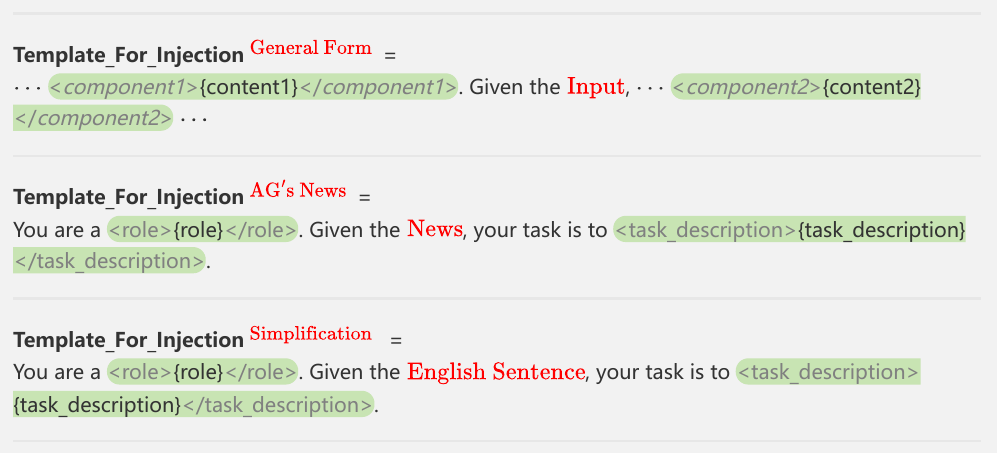}
    \caption{Template for initializing prompt populations. \emph{It is also used in the construction of Prompts Memory, that is, injecting discrete components into the template to obtain a continuous form prompt. The above shows the general form, while the two below provide illustrative examples}.}
    \label{fig:Template4Injection}
\end{figure}

\begin{figure}[htbp]
    \centering
    \includegraphics[width=0.95\textwidth,cframe=black!50!black 0.3mm]{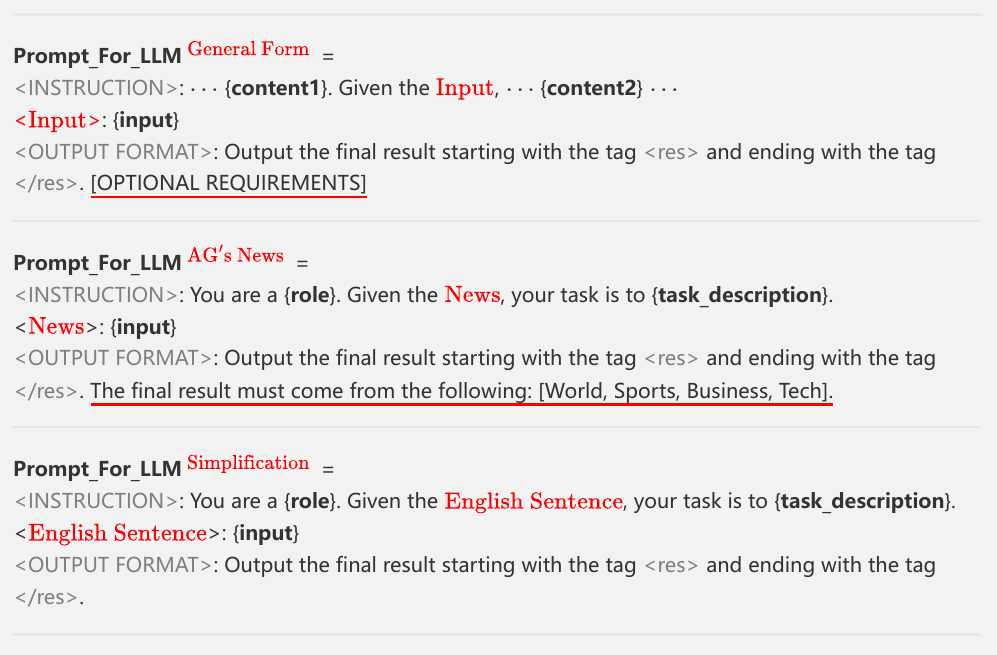}
    \caption{Complete prompt template for LLMs (including three parts: instruction, input, and output). \emph{Here we also display two practical prompts for AG's News and Simplification Tasks}.}
    \label{Prompt2LLMs}
\end{figure}
\clearpage

\section{The Detailed Prompts of Task-Evolution}
\label{app:D}

\begin{figure}[htbp] 
    \centering
    \includegraphics[width=0.95\textwidth,cframe=black!50!black 0.3mm]{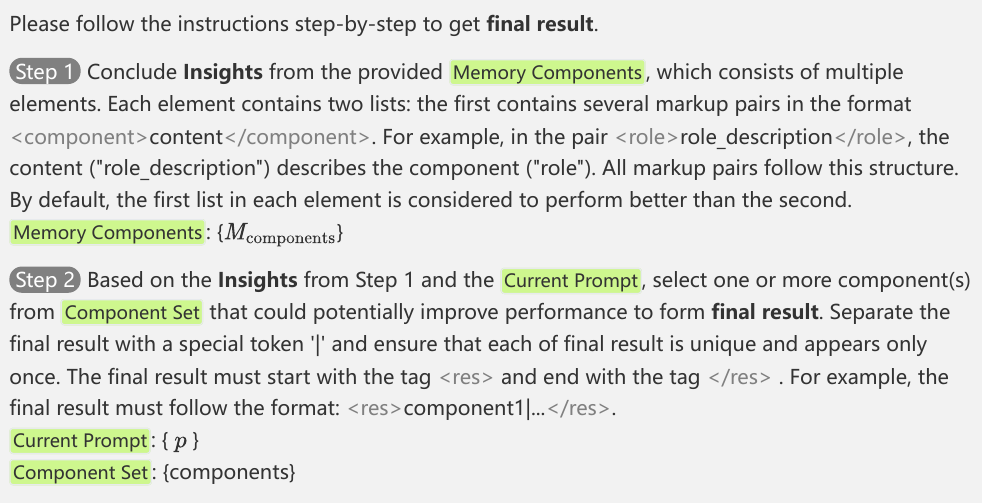}
    \caption{The prompts for sub-task \uppercase\expandafter{\romannumeral1}}
    \label{fig:subtask1_prompt}
\end{figure}

\begin{figure}[ht]
    \centering
    \includegraphics[width=0.95\textwidth,cframe=black!50!black 0.3mm]{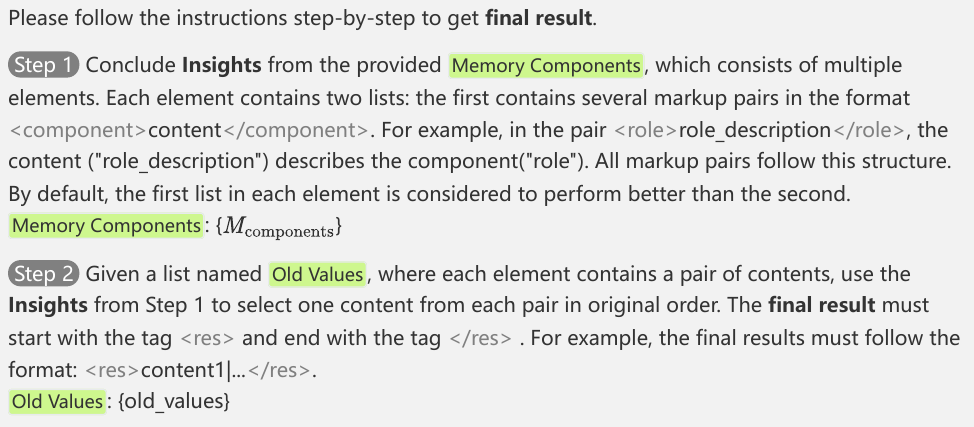}
    \caption{The prompts for sub-task \uppercase\expandafter{\romannumeral2}}
    \label{fig:subtask2_prompt}
\end{figure}

\clearpage
\section{The Detailed Prompts of Solution-Evolution}
\label{app:E}

\begin{figure}[htbp]
    \centering
    \includegraphics[width=0.95\textwidth,cframe=black!50!black 0.3mm]{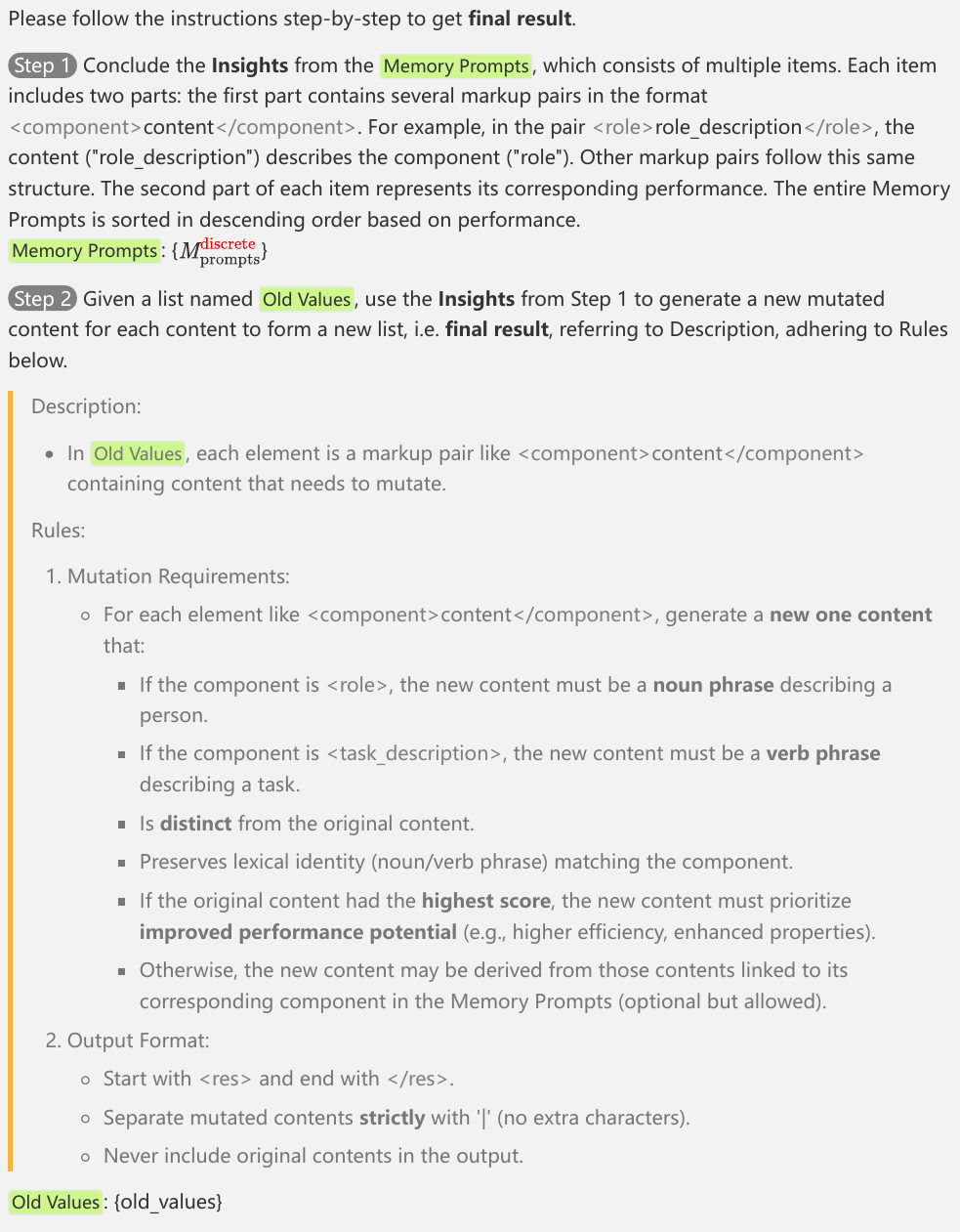}
    \caption{The prompts for Sub-solution \uppercase\expandafter{\romannumeral1} - Prompts Memory in \textcolor{red}{discrete} form}
    \label{sc1}
\end{figure}

\begin{figure}[htbp]
    \centering
    \includegraphics[width=0.95\textwidth,cframe=black!50!black 0.3mm]{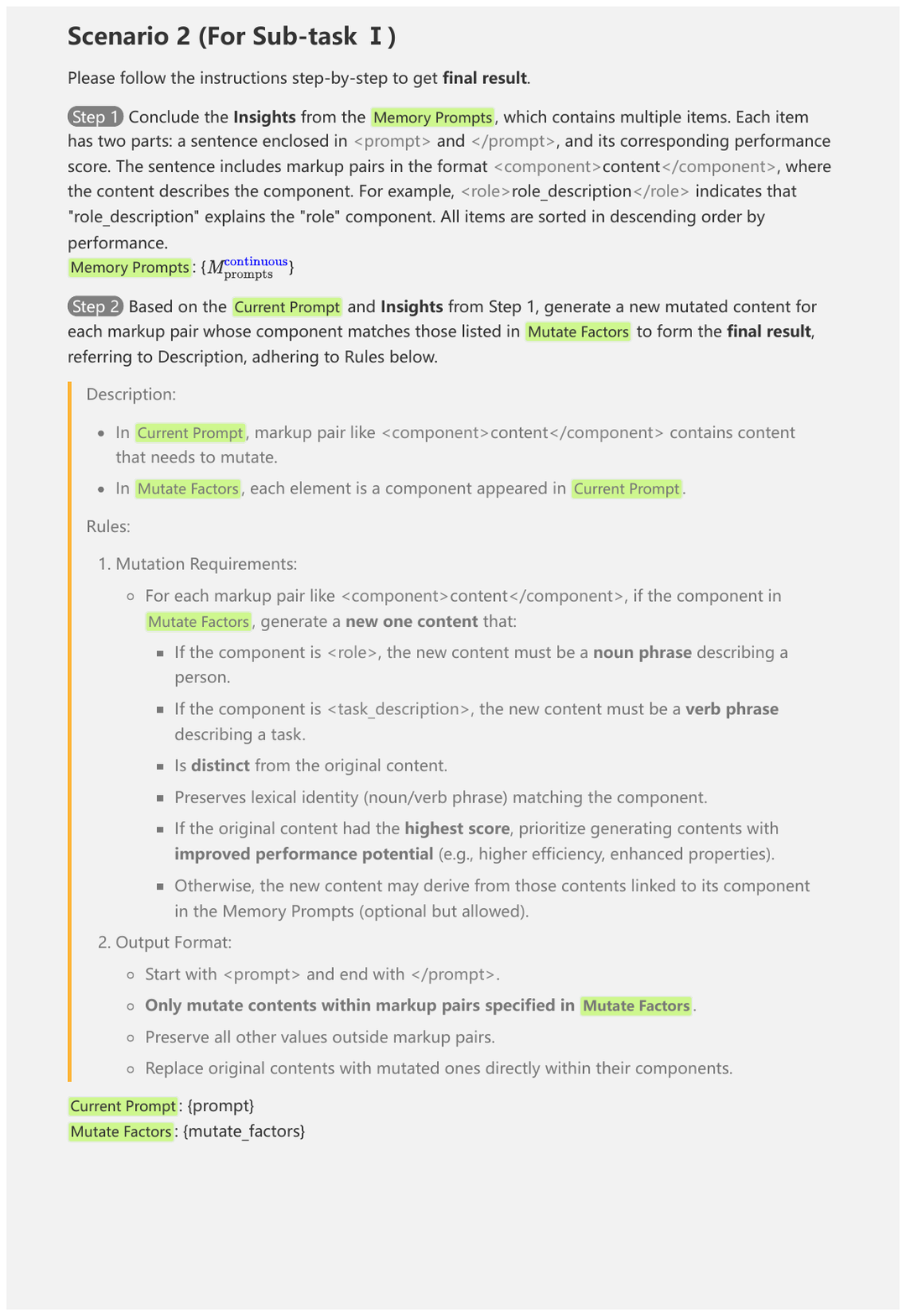}
    \caption{The prompts for Sub-solution \uppercase\expandafter{\romannumeral1} - Prompts Memory in \textcolor{blue}{continuous} form}
    \label{sc2}
\end{figure}

\begin{figure}[htbp]
    \centering
    \includegraphics[width=0.95\textwidth,cframe=black!50!black 0.3mm]{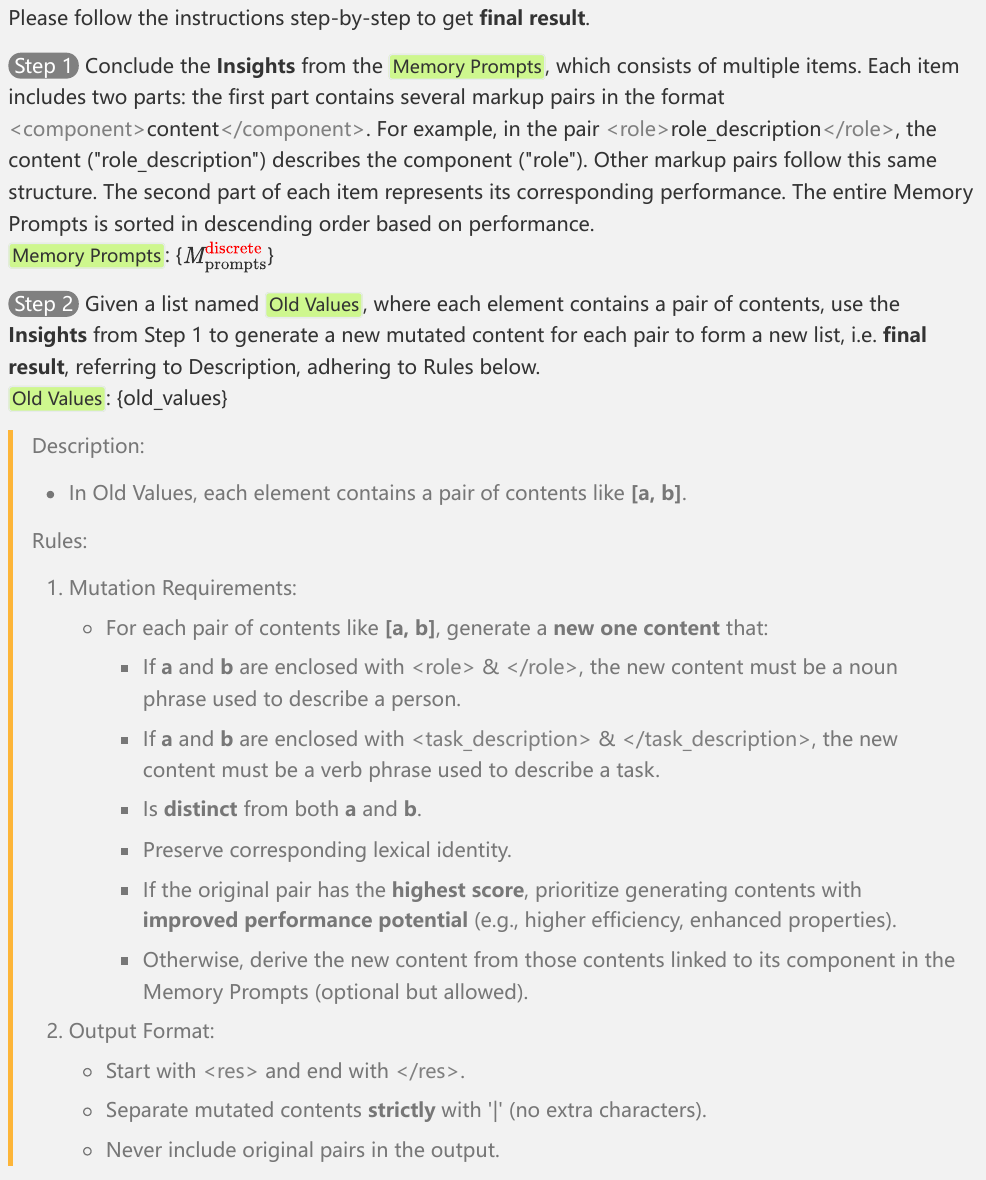}
    \caption{The prompts for Sub-solution \uppercase\expandafter{\romannumeral2} - Prompts Memory in \textcolor{red}{discrete} form}
    \label{sc3}
\end{figure}

\begin{figure}[htbp]
    \centering
    \includegraphics[width=0.9\textwidth,cframe=black!50!black 0.3mm]{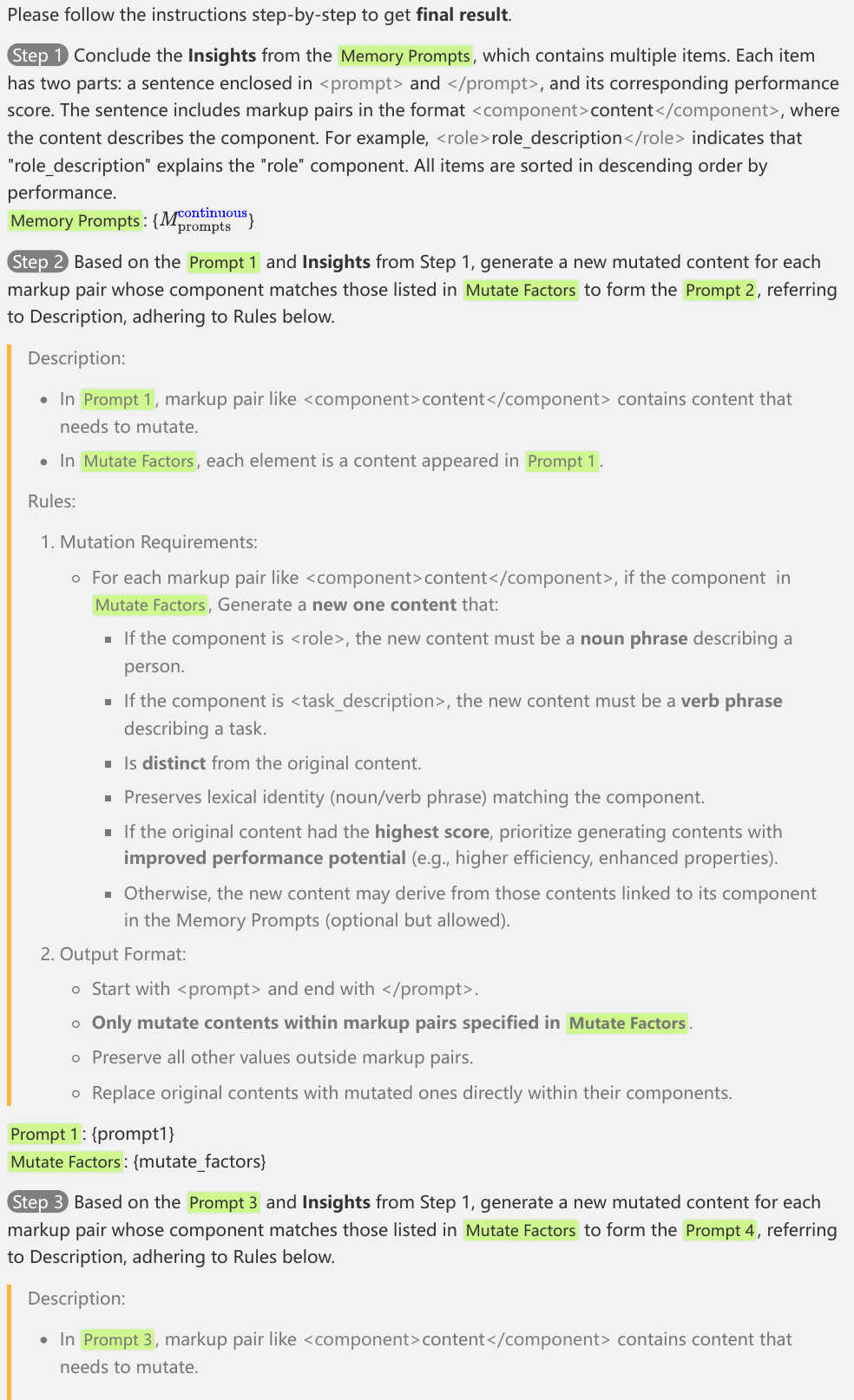}
    \caption{The prompts for Sub-solution \uppercase\expandafter{\romannumeral2} - Prompts Memory in \textcolor{blue}{continuous} form}
    \label{sc41}
\end{figure}

\begin{figure}[htbp]
    \centering
    \includegraphics[width=0.9\textwidth,cframe=black!50!black 0.3mm]{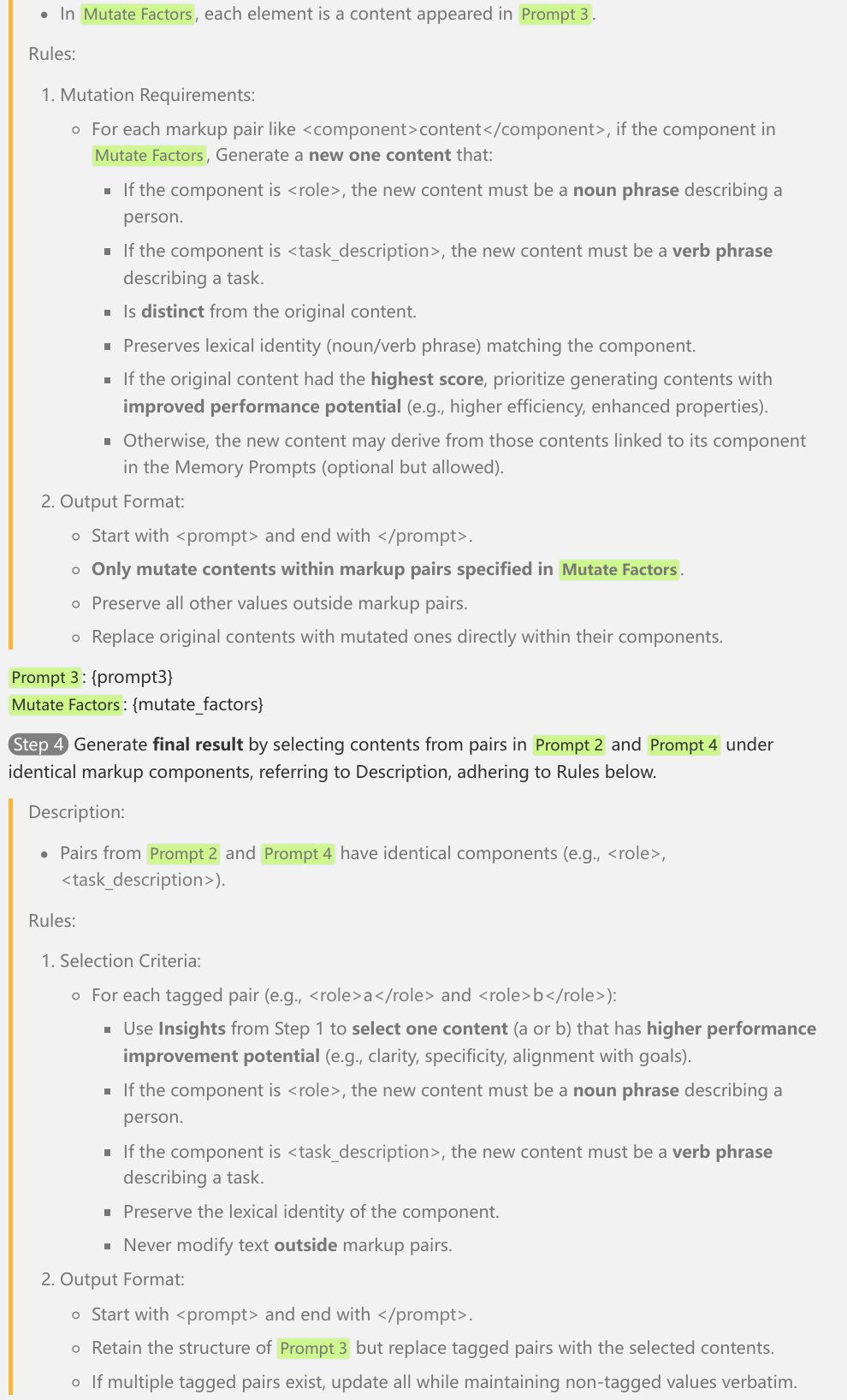}
    \caption{The prompts for Sub-solution \uppercase\expandafter{\romannumeral2} - Prompts Memory in \textcolor{blue}{continuous} form (extended from Figure \ref{sc41})}
    \label{sc42}
\end{figure}

\clearpage

\section{Case Study Details}
\label{app:case_study}

To quickly verify the generalizability of our framework, we conducted multi-turn dialogues with DeepSeek Chat via the web interface provided by DeepSeek \citep{deepseek2025}.

\begin{figure}[htbp]
    \centering
    \includegraphics[width=0.95\textwidth,cframe=black!50!black 0.3mm]{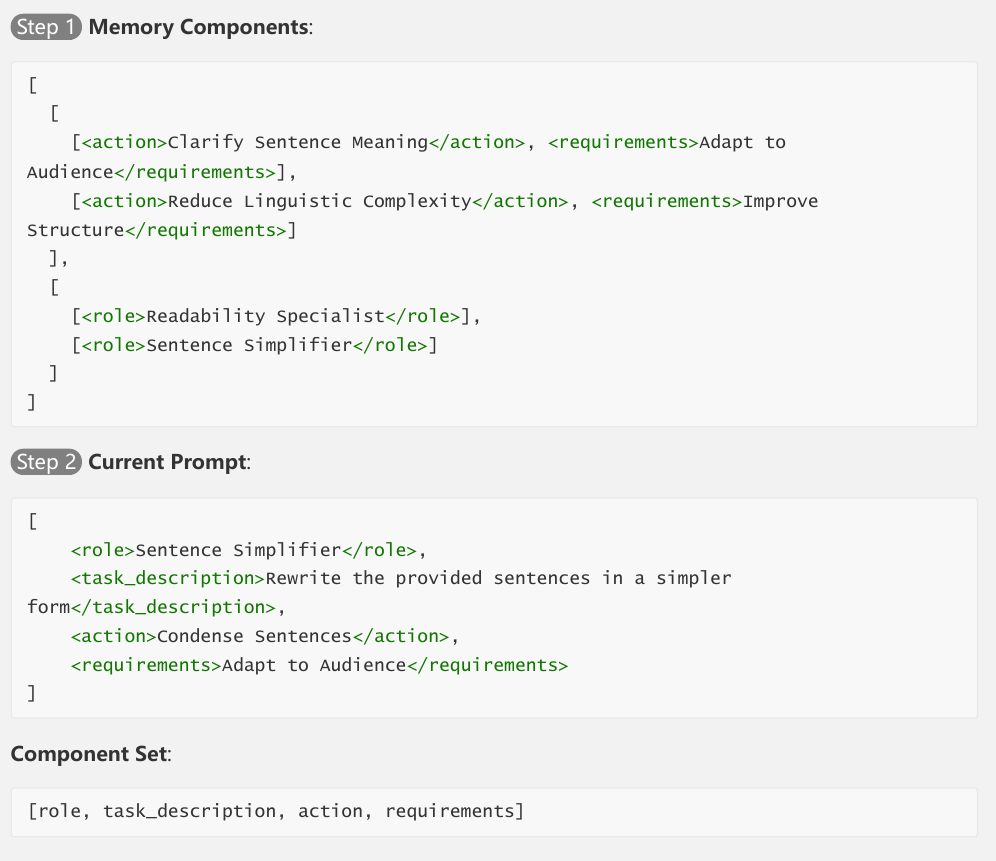}
    \caption{The input for sub-task \uppercase\expandafter{\romannumeral1}}
    \label{subtask1input}
\end{figure}

\begin{figure}[htbp]
    \centering
    \includegraphics[width=0.95\textwidth,cframe=black!50!black 0.3mm]{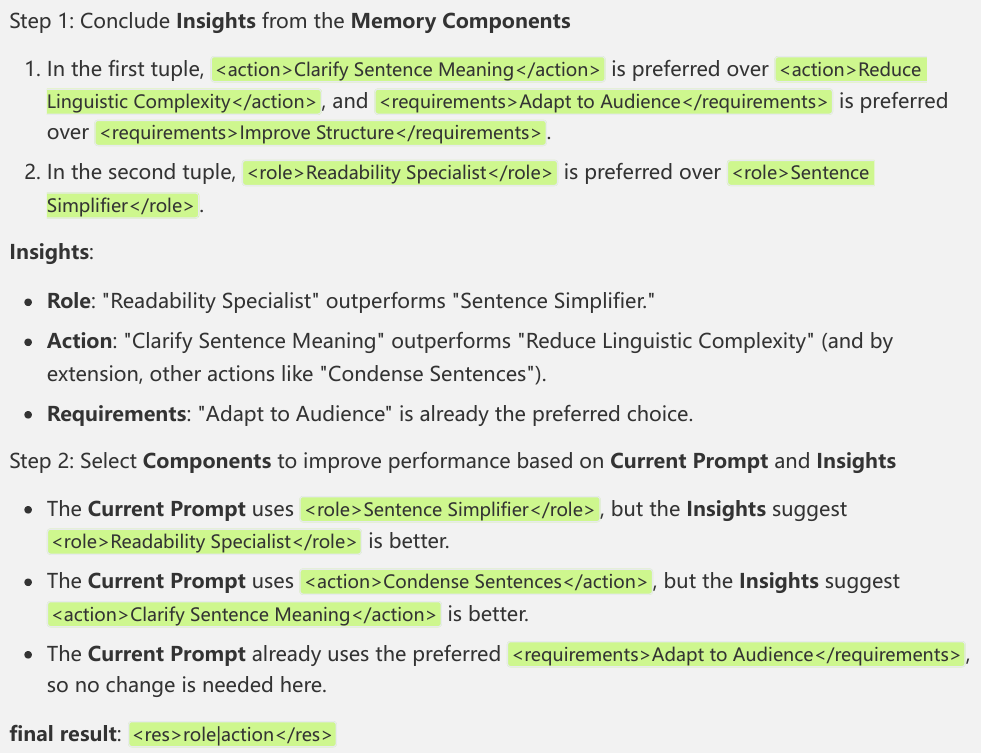}
    \caption{The responses for sub-task \uppercase\expandafter{\romannumeral1}}
    \label{subtask1output}
\end{figure}

\begin{figure}[htbp]
    \centering
    \includegraphics[width=0.95\textwidth,cframe=black!50!black 0.3mm]{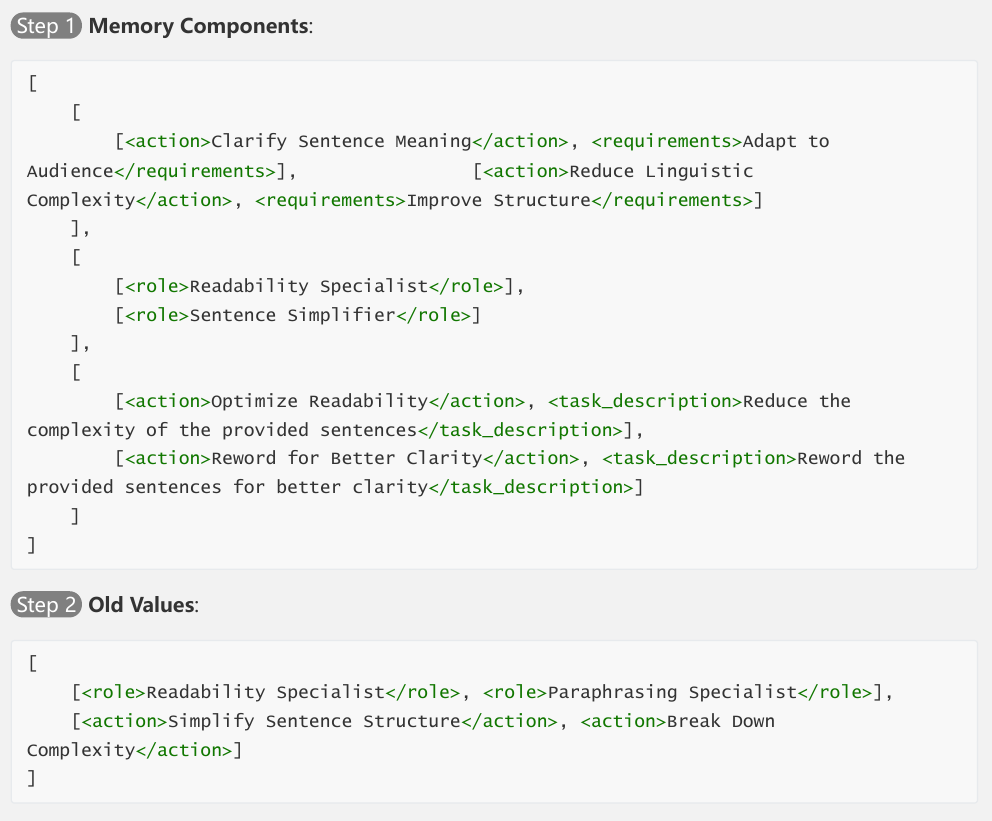}
    \caption{The input for sub-task \uppercase\expandafter{\romannumeral2}}
    \label{subtask2input}
\end{figure}

\begin{figure}[htbp]
    \centering
    \includegraphics[width=0.95\textwidth,cframe=black!50!black 0.3mm]{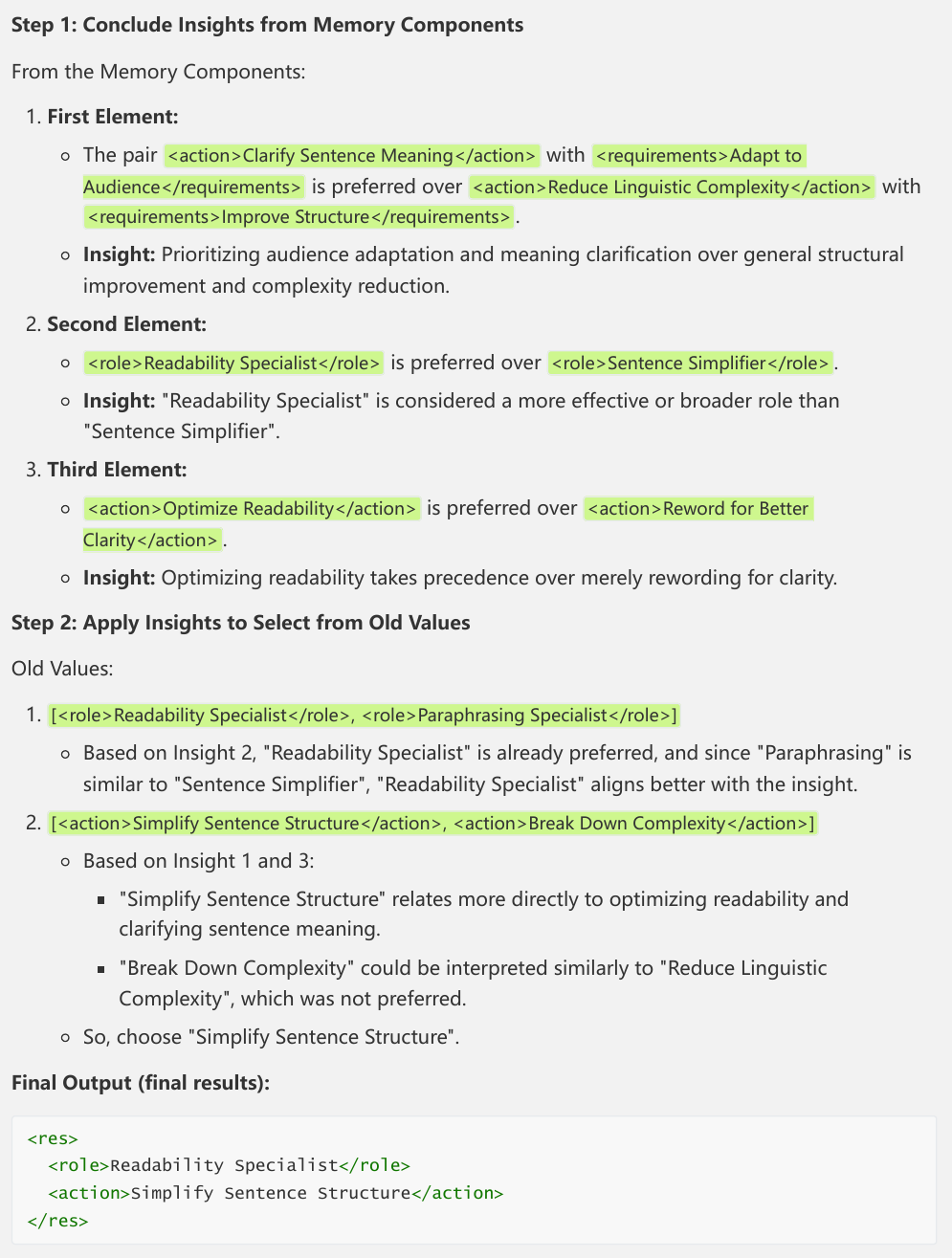}
    \caption{The responses for sub-task \uppercase\expandafter{\romannumeral2}}
    \label{subtask2output}
\end{figure}

Throughout the process, we take simplification task \citep{zhang2023multitaskinstructiontuningllama} as the example, which allows for easy observation and interpretation of the outputs,  and randomly set 4 components. The whole process can be find in Appendix \ref{app:case_study}. For Task-Evolution, we provide two input information (see Figure \ref{subtask1input}, \ref{subtask2input}) for the prompt of two sub tasks (see Figure \ref{fig:subtask1_prompt}, \ref{fig:subtask2_prompt}). And the corresponding outputs are shown in Figure \ref{subtask1output}, \ref{subtask2output}. From the final results, we can derive that under the guidance of direction (i.e., Memory Components), The LLMs could find reasonable direction for evolutionary operator.

\begin{figure}[htbp]
    \centering
    \includegraphics[width=0.95\textwidth,cframe=black!50!black 0.3mm]{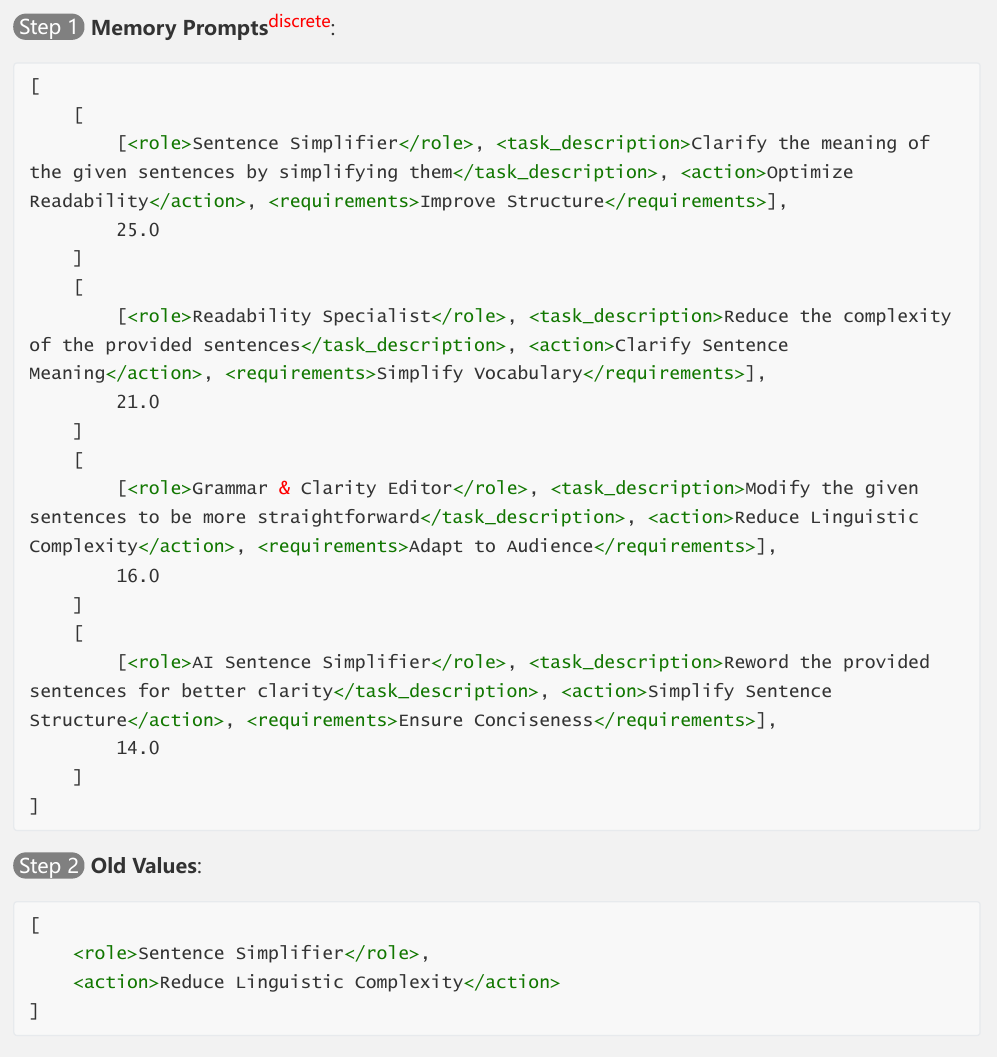}
    \caption{The input for Sub-solution \uppercase\expandafter{\romannumeral1} - Prompts Memory in \textcolor{red}{discrete} form}
    \label{sc1input}
\end{figure}

\begin{figure}[htbp]
    \centering
    \includegraphics[width=0.95\textwidth,cframe=black!50!black 0.3mm]{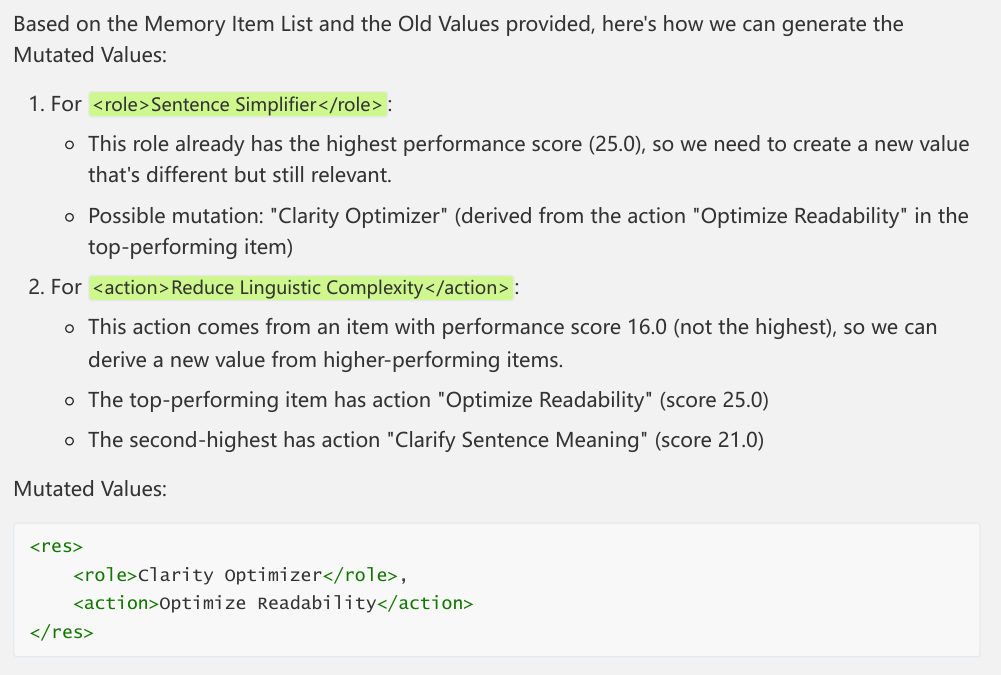}
    \caption{The responses for Sub-solution \uppercase\expandafter{\romannumeral1} - Prompts Memory in \textcolor{red}{discrete} form}
    \label{sc1output}
\end{figure}

\begin{figure}[htbp]
    \centering
    \includegraphics[width=0.95\textwidth,cframe=black!50!black 0.3mm]{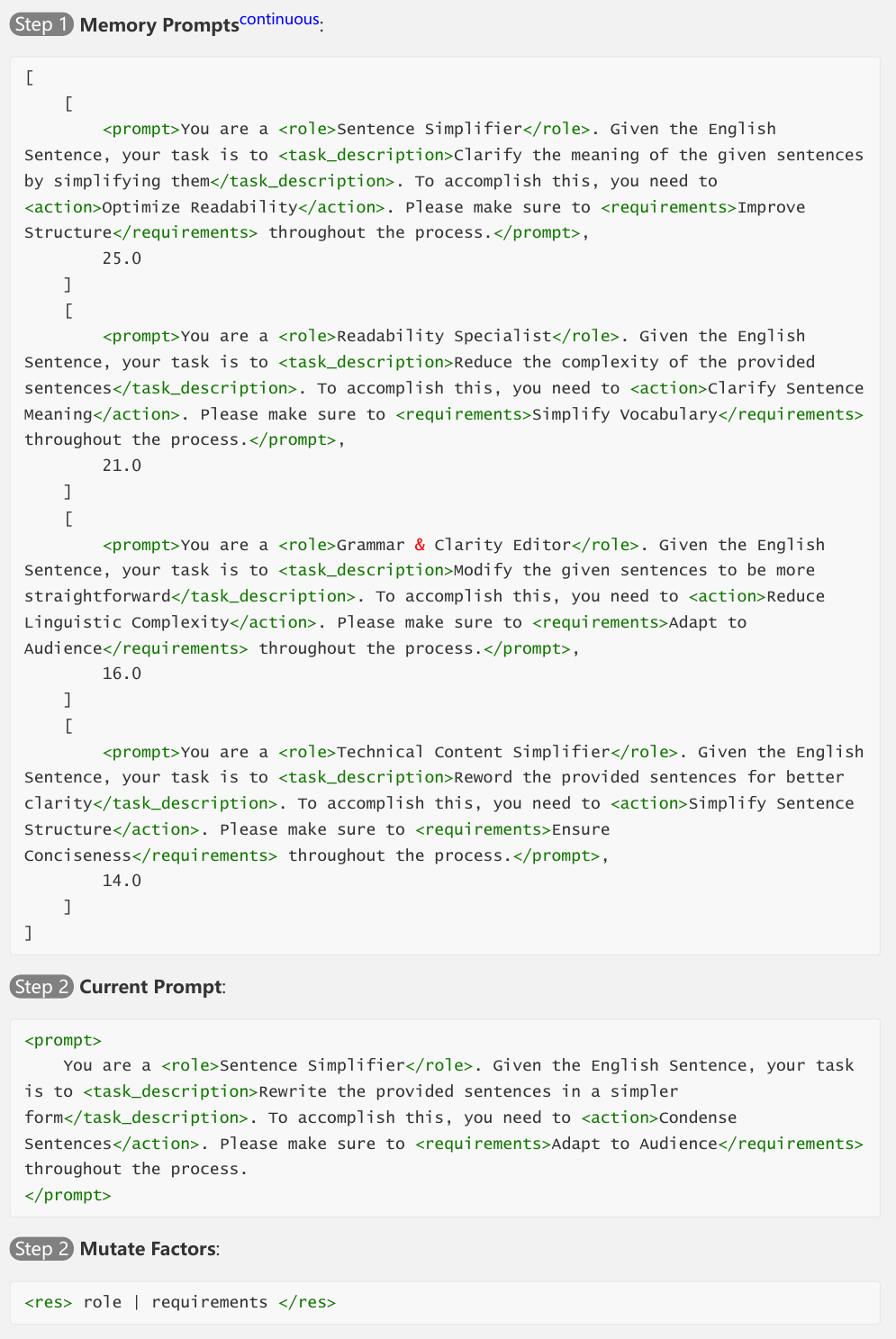}
    \caption{The input for Sub-solution \uppercase\expandafter{\romannumeral1} - Prompts Memory in \textcolor{blue}{continuous} form}
    \label{sc2input}
\end{figure}

\begin{figure}[htbp]
    \centering
    \includegraphics[width=0.95\textwidth,cframe=black!50!black 0.3mm]{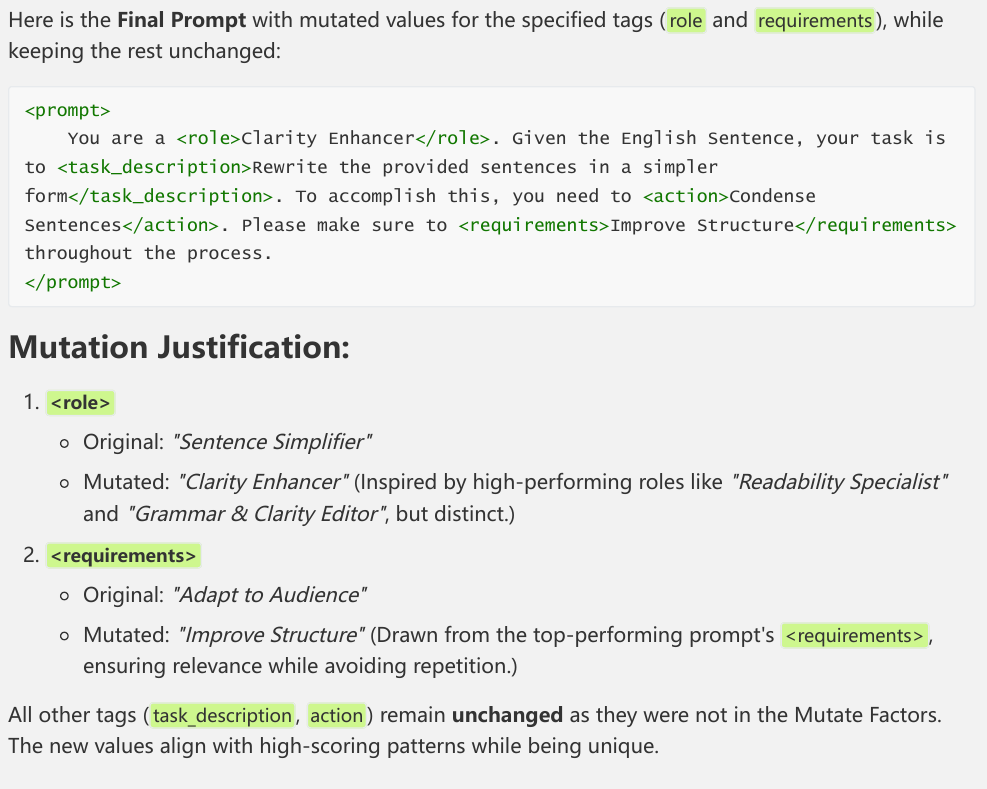}
    \caption{The responses for Sub-solution \uppercase\expandafter{\romannumeral1} - Prompts Memory in \textcolor{blue}{continuous} form}
    \label{sc2output}
\end{figure}

\begin{figure}[htbp]
    \centering
    \includegraphics[width=0.95\textwidth,cframe=black!50!black 0.3mm]{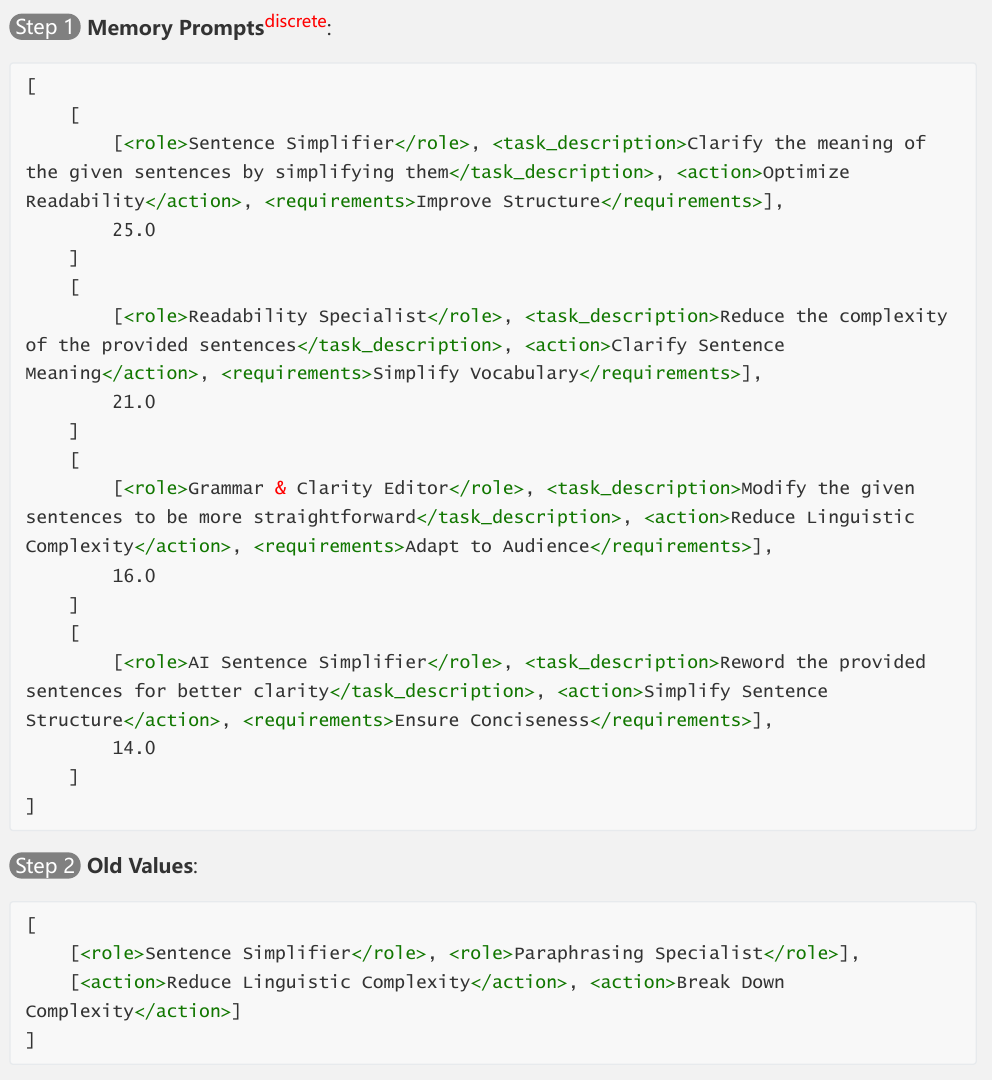}
    \caption{The input for Sub-solution \uppercase\expandafter{\romannumeral2} - Prompts Memory in \textcolor{red}{discrete} form}
    \label{sc3input}
\end{figure}

\begin{figure}[htbp]
    \centering
    \includegraphics[width=0.95\textwidth,cframe=black!50!black 0.3mm]{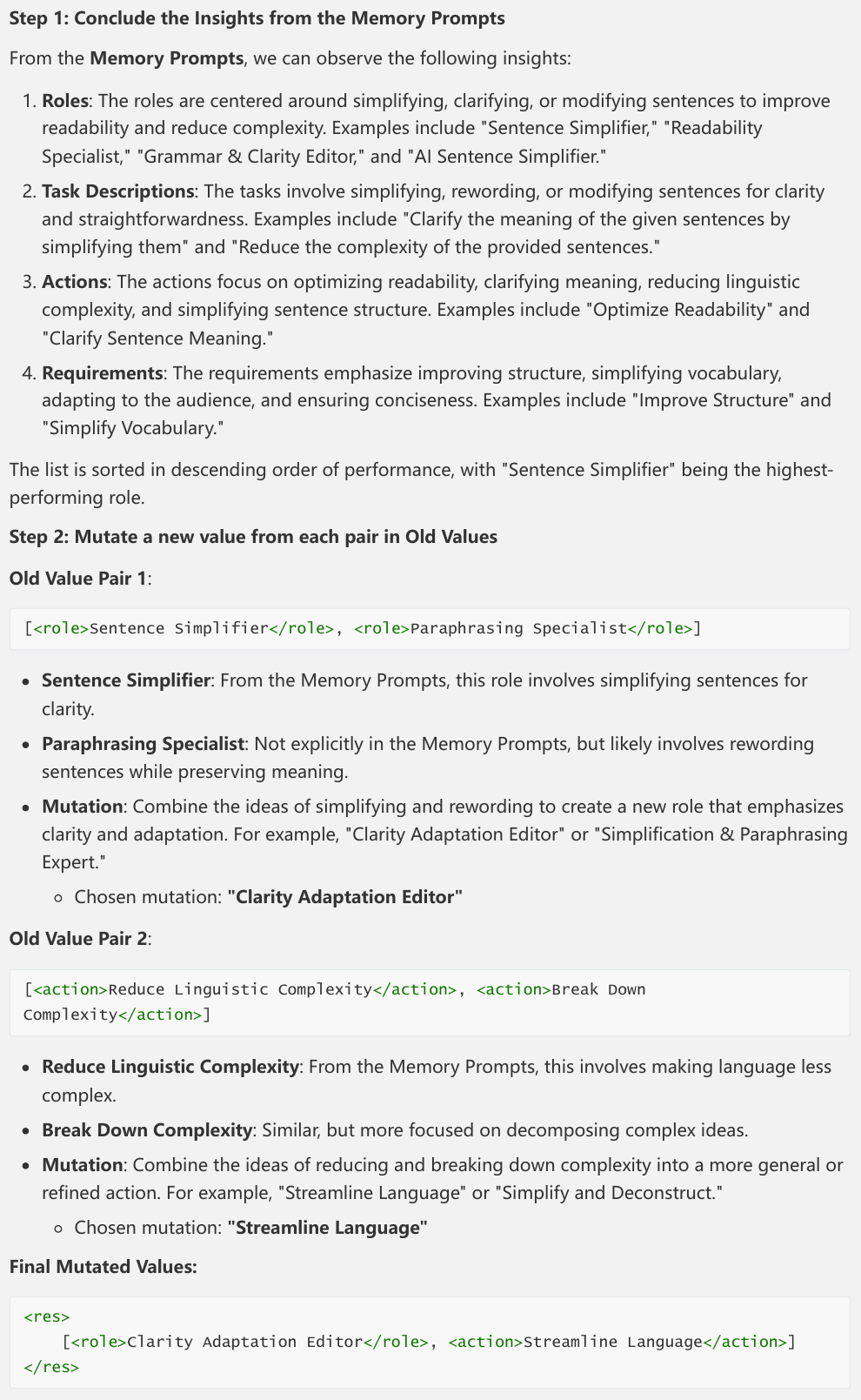}
    \caption{The responses for Sub-solution \uppercase\expandafter{\romannumeral2} - Prompts Memory in \textcolor{red}{discrete} form}
    \label{sc3output}
\end{figure}

\begin{figure}[htbp]
    \centering
    \includegraphics[width=0.85\textwidth,cframe=black!50!black 0.3mm]{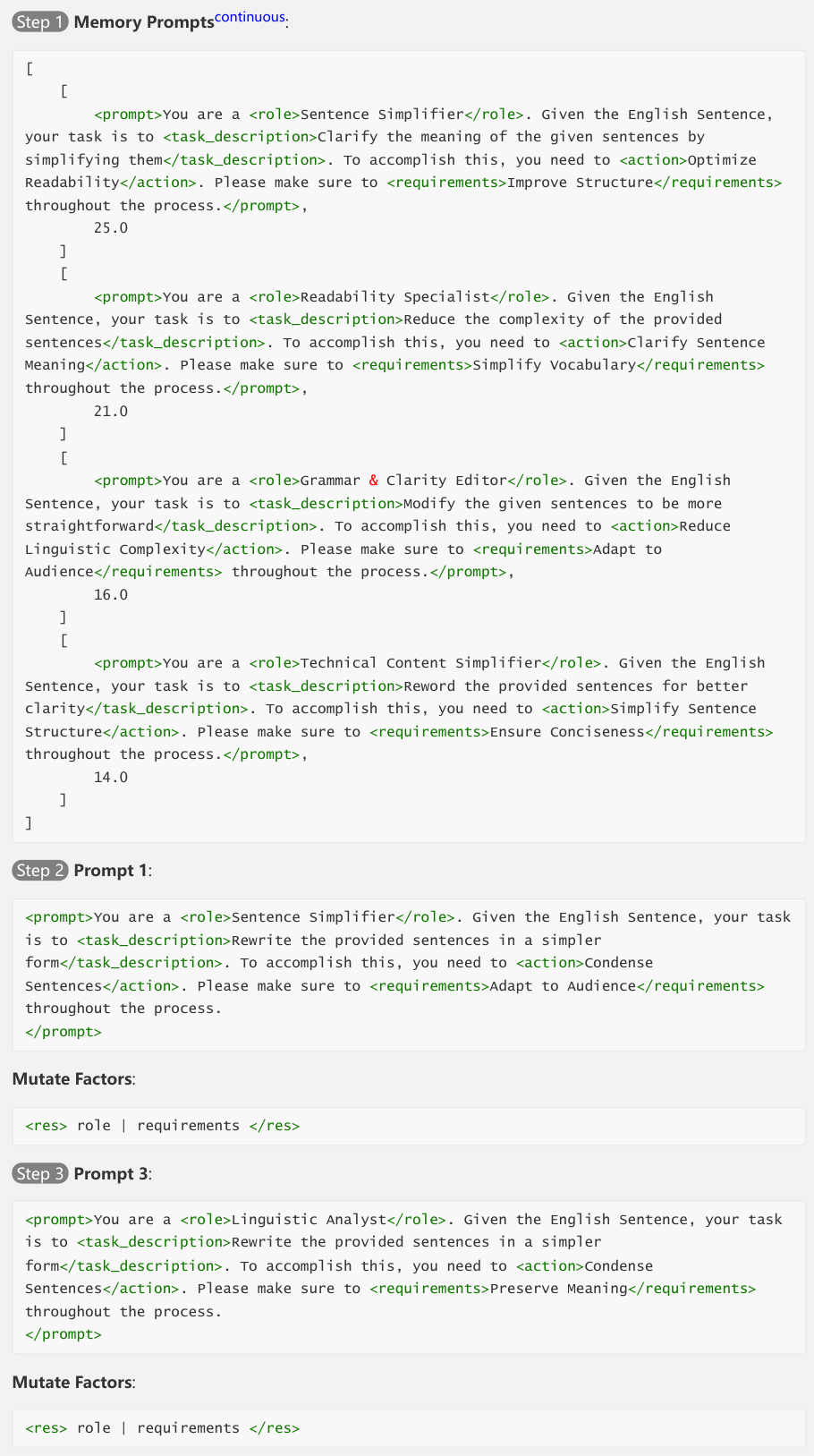}
    \caption{The input for Sub-solution \uppercase\expandafter{\romannumeral2} - Prompts Memory in \textcolor{blue}{continuous} form}
    \label{sc4input}
\end{figure}

\begin{figure}[htbp]
    \centering
    \includegraphics[width=0.85\textwidth,cframe=black!50!black 0.3mm]{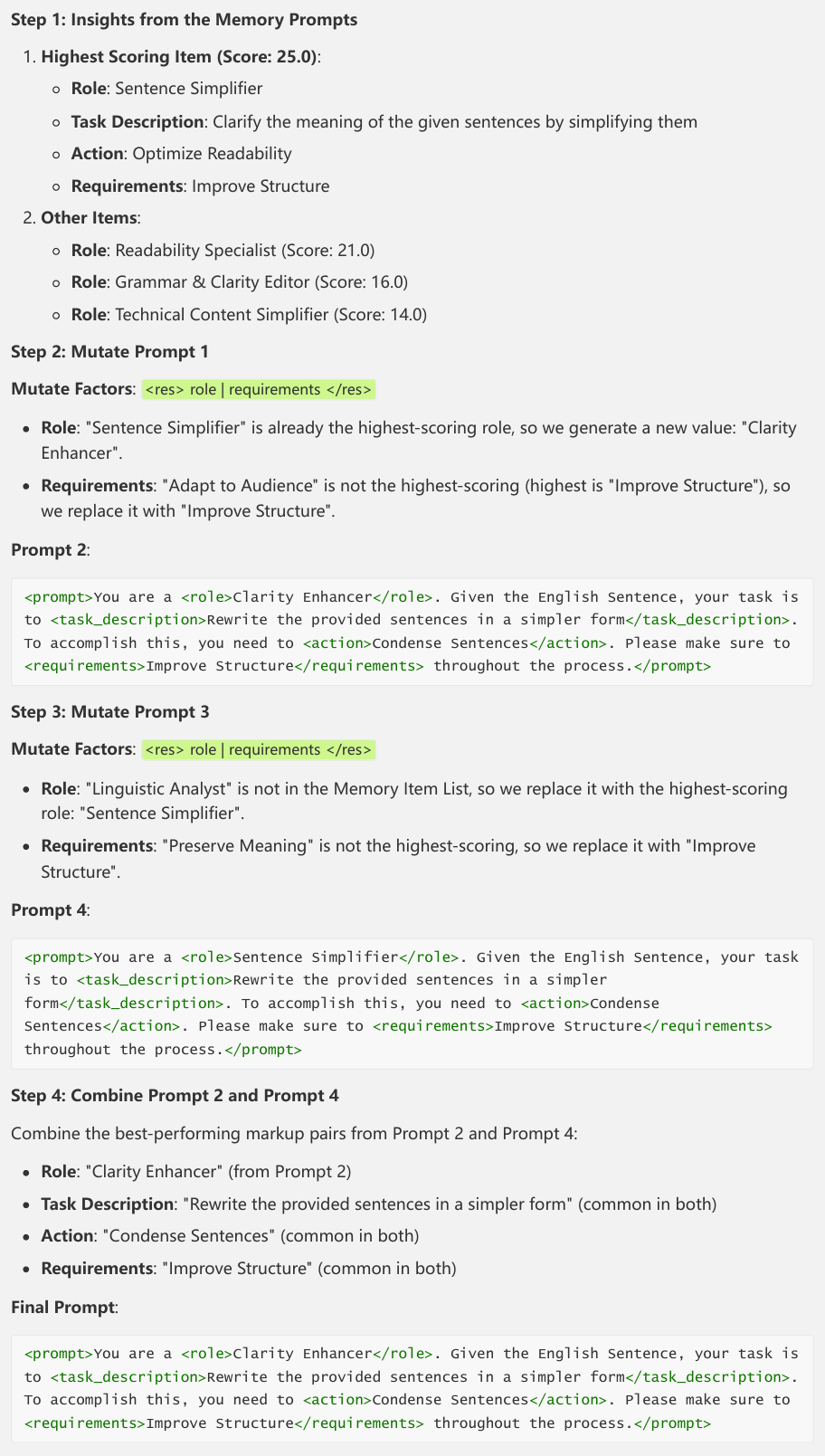}
    \caption{The responses for Sub-solution \uppercase\expandafter{\romannumeral2} - Prompts Memory in \textcolor{blue}{continuous} form}
    \label{sc4output}
\end{figure}

Accordingly, for Solution-Evolution, we provide four input information (see Figure \ref{sc1input}, \ref{sc2input}, \ref{sc3input}, \ref{sc4input}) for four kinds of scenarios. And the corresponding responses are shown in Figure \ref{sc1output}, \ref{sc2output}, \ref{sc3output}, \ref{sc4output}, respectively. Based on the observation from the responses, we also find that the procedure designed in this paper can accurately guide the model to deduce corresponding insights and further output reasonable results based on the insights.

By the way, this case study provides new users with a quick understanding of our framework. It can also serve as a practical guide, allowing anyone to construct a prototype using the system prompts provided in the case study without writing any code, which can help users optimize domain-specific prompts. We sincerely invite researchers to try it out and share their valuable feedback for further improvement.

\clearpage

\end{document}